\documentclass[nohyperref]{article}

\usepackage{microtype}
\usepackage{graphicx}
\usepackage{subfigure}
\usepackage{booktabs} 
\usepackage{algorithm}
\usepackage{algorithmic}

\usepackage{hyperref}

\usepackage[accepted]{icml2022}

\usepackage{amsmath}
\usepackage{amssymb}
\usepackage{mathtools}
\usepackage{amsthm}
\usepackage{bm}

\usepackage[capitalize,noabbrev]{cleveref}

\usepackage{xcolor}
\definecolor{myMagenta}{rgb}{0.9,0,0.4}

\theoremstyle{plain}
\newtheorem{theorem}{Theorem}[section]

\theoremstyle{definition}
\newtheorem{definition}[theorem]{Definition}
\newtheorem{assumption}[theorem]{Assumption}
\theoremstyle{remark}
\newtheorem{remark}[theorem]{Remark}

\usepackage[textsize=tiny]{todonotes}

\icmltitlerunning{Principled Knowledge Extrapolation with GANs}

\begin{document}

\twocolumn[
\icmltitle{Principled Knowledge Extrapolation with GANs}



\icmlsetsymbol{equal}{*}

\begin{icmlauthorlist}
\icmlauthor{Ruili Feng}{ustc}
\icmlauthor{Jie Xiao}{ustc}
\icmlauthor{Kecheng Zheng}{ustc}
\icmlauthor{Deli Zhao}{ant}
\icmlauthor{Jingren Zhou}{ali}
\icmlauthor{Qibin Sun}{ustc}
\icmlauthor{Zheng-Jun Zha}{ustc}

\end{icmlauthorlist}

\icmlaffiliation{ustc}{University of Science and Technology of China, Hefei, China}
\icmlaffiliation{ant}{Ant Research, Hangzhou, China}
\icmlaffiliation{ali}{Alibaba Group, Hangzhou, China}

\icmlcorrespondingauthor{Zheng-Jun Zha}{zhazj@ustc.edu.cn}


\icmlkeywords{Machine Learning, ICML}

\vskip 0.3in
]



\printAffiliationsAndNotice{\icmlEqualContribution} 

\newcommand{\ieno}{\textit{i}.\textit{e}.}
\newcommand{\egno}{\textit{e}.\textit{g}.}
\newcommand{\etcno}{\textit{etc}}
\newcommand{\bphi}{\bm{\phi}}
\newcommand{\btheta}{\bm{\theta}}

\newcommand{\Gtheta}{\bm{G}_{\bm{\theta}}}
\newcommand{\GthetaH}{\bm{G}_{\bm{\theta}^H}}

\newcommand{\GthetaX}{\bm{G}_{\bm{\theta}^{\mathcal{X}}}}

\newcommand{\Dphi}{\bm{D}_{\bm{\phi}}}
\newcommand{\z}{\bm{z}}
\newcommand{\x}{\bm{x}}
\newcommand{\w}{\bm{w}}

\begin{abstract}
Human can extrapolate well, generalize daily knowledge into unseen scenarios, raise and answer counterfactual questions. To imitate this ability via generative models, previous works have extensively studied explicitly encoding Structural Causal Models (SCMs) into architectures of generator networks. This methodology, however, limits the flexibility of the generator as they must be carefully crafted to follow the causal graph, and demands a ground truth SCM with strong ignorability assumption as prior, which is a nontrivial assumption in many real scenarios. Thus, many current causal GAN methods fail to generate high fidelity counterfactual results as they cannot easily leverage state-of-the-art generative models. In this paper, we propose to study counterfactual synthesis from a new perspective of knowledge extrapolation, where a given knowledge dimension of the data distribution is extrapolated, but the remaining knowledge is kept indistinguishable from the original distribution. We show that an adversarial game with a closed-form discriminator can be used to address the knowledge extrapolation problem, and a novel principal knowledge descent method can efficiently estimate the extrapolated distribution through the adversarial game. Our method enjoys both elegant theoretical guarantees and superior performance in many scenarios.
\end{abstract}
\definecolor{cf}{RGB}{46, 117, 182}
\begin{figure}[t]
    \centering
    \includegraphics[width=1.0\linewidth]{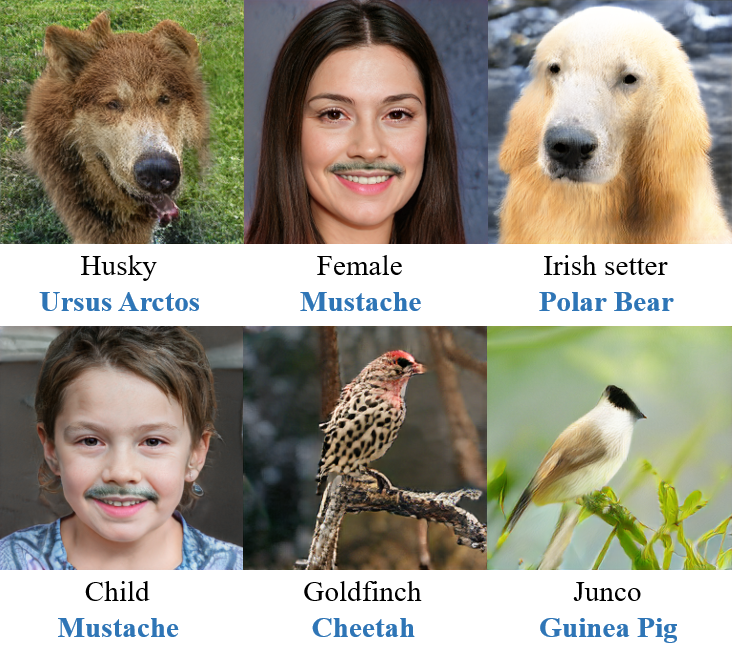}\vspace{-0.4cm}
    \caption{Knowledge extrapolation. All the above objects are counterfactual, rarely existent in real world. The origin domains of them are written benzene, and the extrapolated knowledge is marked in \textcolor{cf}{\textbf{purple}}.}\vspace{-0.4cm}
    \label{fig:teaser}
\end{figure}

\section{Introduction}
Human beings exhibit remarkable ability of cognitive extrapolation~\cite{ehrlich2005human,beck2006children} in a variety of aspects. For example, we can accurately extrapolate the motion of objects~\cite{ehrlich2005human}, imagine unseen objects~\cite{kocaoglu2018causalgan}, raise and answer counterfactual questions~\cite{beck2006children}. There is a temptation to wonder whether Generative Adversarial Networks (GANs) \cite{goodfellow2014generative} can generalize well to examples whose distribution is arbitrarily far from that of the given training data (or even counterfactual examples), as shown in Fig. \ref{fig:teaser}. Specifically, this knowledge extrapolation ability of GANs can be 
reflected as synthesizing out-of-distribution examples and manipulating semantic features to constitute counterfactual combinations.

Counterfactual synthesis \cite{kocaoglu2018causalgan, sauer2020counterfactual, nemirovsky2020countergan, yang2021causalvae, averitt2020counterfactual, thiagarajan2021designing} is one of the most promising tasks to achieve the general goal of knowledge extrapolation in GANs. For counterfactual synthesis, if brusquely ignoring differences in details, most existing methods follow the framework of the same fashion --- directly modeling a Structural Causal Model (SCM) \cite{pearl2009causal} in well-designed architectures of generator networks. Factors of interest in the causal graph are designed as labels to control the synthesis of the generator. However, this type of approaches is somewhat inflexible. Specifically, it demands a prior SCM to identify all the causalities in the data generation process. Theoretically, constructing a prior SCM with photo-realistic synthesis effect can be embarrassingly difficult  if many potential factors and obscure entanglements are involved \cite{pearl2009causal, sekhon2008neyman, holland1986statistics}. In addition, the rigorous inference of causal effects  needs the strong assumptions like `strong ignorability' (\ieno, assuming no unobserved confounders) \cite{sekhon2008neyman, holland1986statistics}, which is hard to verify in practice \cite{pearl2009causality,pearl2009causal}. Therefore, these methods are limited in many knowledge extrapolation scenarios because it is inconvenient to deduce a prior SCM. In addition to the demand of prior SCMs, these approaches also put constraints on the network design that has to be consistent with the prior SCM. Concretely, they need to construct generators that each network component corresponds to a causal graph factor to yield causal interventions \cite{pearl2009causal,holland1986statistics}. While new GAN architectures adapt rapidly, it is of great challenge to directly apply those methods to state-of-the-art GANs (\egno, StyleGANs \cite{karras2019style,karras2020analyzing} or BigGAN \cite{brock2018large}), thus limiting their overall performance regarding high-fidelity counterfactual synthesis.

In this paper, we propose a principled knowledge extrapolation method to circumvent the two embarrassing flaws of current methods, and provide a new viewpoint for knowledge extrapolation of GANs. Instead of modeling prior causal graphs in generators, here we turn to a simple hypothesis that the original distribution and extrapolated distribution are indistinguishable, except for the extrapolated knowledge dimension. For example, if there is a hypothetical distribution in which all people (including women and children) could wear beards, then the information to distinguish this hypothetical distribution from real world data only exists in the distribution of beards. Thus the beard-irrelevant information (\ieno, gender, age or other knowledge dimensions) cannot contribute to distinguish these two distributions, as the remaining knowledge is the same between these two distributions.
Under this assumption, we prove that an adversarial learning strategy \cite{goodfellow2014generative} can approximate the hypothetical distribution with a closed-form discriminator when we manage to preserve the irrelevant knowledge unchanged during the adversarial game. To achieve this goal, a novel Principal Knowledge Descent (PKD) method is proposed to solve a sparse paradigm in the parameter space that starts from the pretrained generator distribution to an approximation of the hypothetical distribution. The sparse paradigm will involve only the most related parameters to our knowledge of interest, thus posing negligible influence to the other knowledge.

In conclusion, the contributions of this paper include:
\begin{itemize}
    \item We propose a new principled GAN knowledge extrapolation method that is flexible to use and can be easily adapted to state-of-the-art GAN architectures;
    \item We design a novel sparse descent strategy to efficiently estimate the extrapolated distribution based on our theory;
    \item The proposed method is the first to successfully synthesize high-fidelity counterfactual results in various image data domains.
\end{itemize}

\section{Related Work}
The prevalence of GAN \cite{goodfellow2014generative}  has aroused researchers' ambitions to utilize various novel GANs to synthesize counterfactual data under the guidance of prior structural causal models. CausalGAN \cite{kocaoglu2018causalgan} proposes to learn a causal implicit model through adversarial training with a given causal graph for facial attribute disentanglement. CounterGAN \cite{nemirovsky2020countergan} employs a residual generator to improve counterfactual realism and actionability compared to regular GANs. Counterfactual Generative Network (CGN) \cite{sauer2020counterfactual} suggests to decouple the ImageNet generation into four aspects of the shape, texture, background, and composer. CGN explicitly models the causality in the four aspects to yield counterfactual combinations of them, like triumphal arch with the elephant texture. CausalVAE \cite{yang2021causalvae} employs structural causal layer to encode prior causalities. \citeauthor{thiagarajan2021designing}  exploits deep image priors from a U-Net \cite{ronneberger2015u} and a classification model to synthesize counterfactual images. Those methods provide compelling insights to the causal explanation of black-box generative models, but put extra limitations on generator architectures, thus generally yielding much less plausible synthesis than state-of-the-art GAN models. Also, these methods rely on prior causal models, which grossly limit their generalization to other GAN architectures.

\section{Method}
Given a data domain $\mathcal{X}$ and a data distribution $\mathbb{P}_{\mathcal{X}}$, we assume that there is already a pretrained generator network $\GthetaX:\mathcal{Z}\rightarrow\mathcal{X}$ that captures the data distribution, and a posterior probability $\mathbb{P}_l(\x)=\mathbb{P}(l|\x)$ \footnote{This paper uses $\mathbb{P}$ to denote the probability density.} for a knowledge of interest $l$. The pretrained generator network transports the prior distribution $\mathbb{P}_{\mathcal{Z}}$ (which is usually the standard Gaussian) on the latent space $\mathcal{Z}$ into the data distribution $\mathbb{P}_{\mathcal{X}}$, \ieno, $\mathbb{P}_{\GthetaX}=\mathbb{P}_{\mathcal{X}}$ \cite{goodfellow2014generative}, with $\btheta^{\mathcal{X}}$ being its parameters at convergence. The generator can be obtained from a pretrained GAN \cite{goodfellow2014generative}, VAE \cite{kingma2013auto}, or other smooth parametric methods that yield generative components \cite{kingma2018glow, dinh2016density}. The posterior probability $\mathbb{P}_l$ can be obtained through classification or regression neural networks on knowledge $l$, or other smooth parametric methods that yield posterior estimation of $l$. For example, a most common case of counterfactual synthesis is a GAN that generates facial images, together with classifiers that identify the posteriors of semantic attributes such as `mustache', `age', `gender', etc. \cite{kocaoglu2018causalgan}.

\begin{figure}[t]
    \centering
    \includegraphics[width=1.0\linewidth]{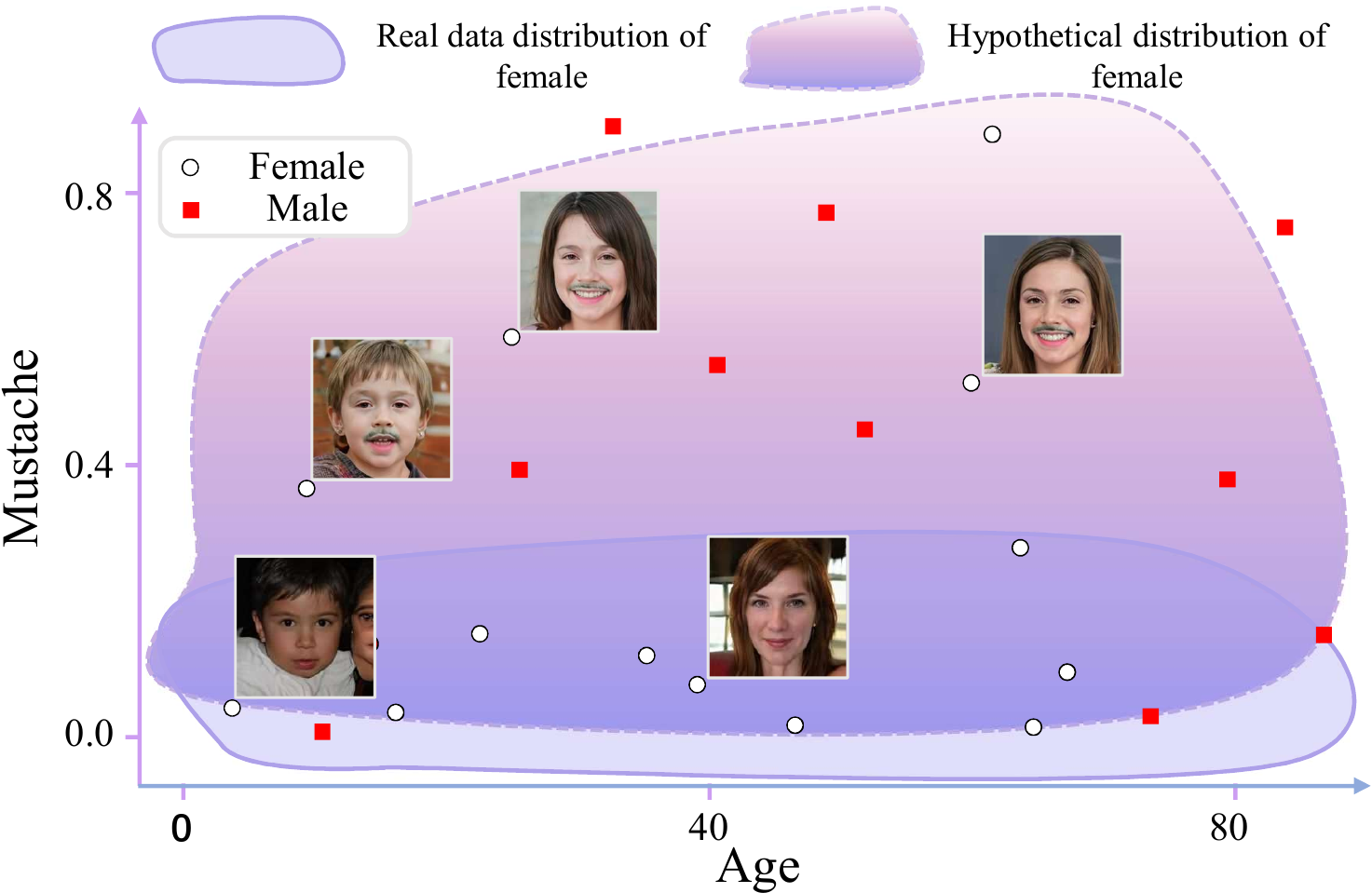}\vspace{-0.4cm}
    \caption{Illustration of the hypothetical distribution. Real data distribution excludes the cases of children or female in mustache. The hypothetical distribution extrapolates to those counterfactual cases, but keeps the other aspects unchanged, especially the confounder factors (factors that influence both the dependent variable (face image) and independent variable (mustache), causing a spurious association) like gender or age of the faces.}
    \label{fig:moti}\vspace{-0.4cm}
\end{figure}

Our task here is to infer a hypothetical distribution $\mathbb{P}_H$, where it only differs from the real data in the knowledge of interest, and is indistinguishable from the real data distribution $\mathbb{P}_{\mathcal{X}}$ among all the remaining knowledge. To capture this hypothetical distribution, we want to get the parametric value $\btheta^H$ such that $\mathbb{P}_{\GthetaH}=\mathbb{P}_H$. Fig. \ref{fig:moti} illustrates the example of `mustache'. If the real data are facial images and the knowledge of interest is `mustache', then we want to obtain a generator that is capable of synthesizing images where all people, including women and children, can wear the `mustache' while all the other semantic knowledge is maintained realistic and plausible. The hypothetical distribution extrapolates the real data distribution along the dimension of knowledge of interest and is counterfactual in the real world. Here the key difference between our method and previous methods is that we bypass  SCM to directly yield counterfactual synthesis results with given pretrained generators, and eliminate the limitation to generator architectures so that we can directly apply our method to any given generative models. 

Here we propose a different perspective to solve the hypothetical distribution $\mathbb{P}_H$ from adversarial training with GANs. Theoretically, the adversarial training of GAN models will terminate at the global optimum of the generated distribution equal to the data distribution \cite{goodfellow2014generative}. Thus, if we aim to alter the data distribution to the hypothetical distribution, we may formulate an adversarial game that will halt at the generated distribution equal to the hypothetical distribution. So we propose to solve 
\begin{equation}\label{eq:hypo_gan}
    \begin{aligned}
        \min_{\Gtheta}\max_{\Dphi} &V(\Dphi,\Gtheta)=\mathbb{E}_{\x\sim\mathbb{P}_{H}}\left[\log(\Dphi(\x))\right]\\
        &+\mathbb{E}_{\x\sim\mathbb{P}_{\Gtheta}}\left[\log(1-\Dphi(\x))\right],
    \end{aligned}
\end{equation}
where $\btheta$ and $\bphi$ are parameters, and $\Dphi$ is the discriminator network. The problem here is that $\mathbb{P}_{H}$ is merely hypothetical, meaning that we do not have any sample from it at hand, so evaluating the value of the term $\mathbb{E}_{\x\sim\mathbb{P}_H}\left[\log(\Dphi(\x))\right]$ of $V(\Dphi,\Gtheta)$, or gradients of it seems to be impossible with typical Monte Carlo methods \cite{hammersley2013monte}. In the next section, a novel adversarial extrapolation in the knowledge dimension will be deduced to address this problem.

\subsection{Adversarial Extrapolation in Indiscernibility Space}
The typical algorithm to solve Problem (\ref{eq:hypo_gan}) is to alternately optimize the generator and discriminator to attain a Nash equilibrium \cite{goodfellow2014generative}. For the hypothetical distribution, however, we are allowed to simplify the training considerably with an indistinguishable assumption.
\begin{assumption}[Indistinguishable assumption]\label{as:ind}
The real data distribution $\mathbb{P}_{\mathcal{X}}$ is indistinguishable from the hypothetical distribution $\mathbb{P}_H$ except for the altered knowledge $l$. 
\end{assumption}
\begin{definition}[Indiscernibility Space]
We denote the collection of all the parameters that can induce generated distribution satisfying Assumption \ref{as:ind} as the indiscernibility space $\mathcal{I}^l$ of knowledge $l$, \ieno,
    $\mathcal{I}^l=\{\btheta:\mathbb{P}_{\Gtheta}$ \textit{is indistinguishable from the hypothetical distribution} $\mathbb{P}_{H}$
    \textit{except for the knowledge of interest} $l\}$.
\end{definition}

Apparently, we have $\btheta^{\mathcal{X}},\btheta^H\in\mathcal{I}^l$, and $\btheta^H$ is the global optimum of Problem (\ref{eq:hypo_gan}). Thus, solving Problem (\ref{eq:hypo_gan}) is equivalent to solving it inside the indiscernibility space $\mathcal{I}^l$:
\begin{equation}\label{eq:hypo_con_gan}\vspace{-0.1cm}
    \begin{aligned}
        \min_{\Gtheta\in\mathcal{I}^l}\max_{\Dphi}&V(\Dphi,\Gtheta)=\mathbb{E}_{\x\sim\mathbb{P}_{H}}\left[\log(\Dphi(\x))\right]\\
        &+\mathbb{E}_{\x\sim\mathbb{P}_{\Gtheta}}\left[\log(1-\Dphi(\x))\right].
    \end{aligned}\vspace{-0.2cm}
\end{equation}

\paragraph{Investigate Assumption \ref{as:ind} from knowledge $l$.} We now consider to solve the the optimal discriminator for generators inside the Indiscernibility Space of Problem (\ref{eq:hypo_con_gan}). Assume $\btheta\in\mathcal{I}^l$. 
Distinguishing which distribution between $\mathbb{P}_{\Gtheta}$ and $\mathbb{P}_{\mathcal{X}}$ a sample $\x$ is more likely to be sampled from can leverage the ratio $\frac{\mathbb{P}_{H}(\x)}{\mathbb{P}_{\Gtheta}(\x)}$. If this ratio is larger than one, then $\x$ is more likely from $\mathbb{P}_{\mathcal{X}}$, otherwise from $\mathbb{P}_{H}$. As $\mathbb{P}_{\Gtheta}$ is indistinguishable from $\mathbb{P}_H$ except for $l$, this ratio should be purely decided by the posterior distribution of knowledge $l$ on sample $\x$. Namely, there is some posterior distribution $\mathbb{P}_l(\x)=\mathbb{P}(l|\x)$ of knowledge $l$, such that
\begin{equation}\label{eq:indistin}\vspace{-0.1cm}
    \frac{\mathbb{P}_{H}(\x)}{\mathbb{P}_{\Gtheta}(\x)}=\frac{\mathbb{P}_l(\x)}{1-\mathbb{P}_l(\x)}.
\end{equation}

\begin{remark}
Eq. (\ref{eq:indistin}) means that distinguishing $\x$ amounts to distinguishing knowledge $l$. When sampling from $\mathbb{P}_{H}$, knowledge $l$ is more likely to appear. While sampling from $\mathbb{P}_{\Gtheta}$, knowledge $l$ is more likely to be ignored. Thus, $\mathbb{P}_H$ also extrapolates knowledge $l$ beyond the original distribution of $\mathbb{P}_{\Gtheta}$.
\end{remark}

An essential property of the indiscernibility space then immediately follows, that when constraining the generator in the indiscernibility space, the optimal discriminator admits a closed-form solution. Rigorously, we have the following:
\begin{theorem}\label{th:optD}
If $\Gtheta\in\mathcal{I}^l$, then the optimal discriminator of problem $\max_{\Dphi}V(\Dphi,\Gtheta)$ is $\bm{D}_{\bphi^*}(\x)=\mathbb{P}_l(\x)$ for some probability distribution $\mathbb{P}_l$ of knowledge $l$. Thus, Problem (\ref{eq:hypo_con_gan}) is equivalent to
\begin{equation}\label{eq:hypo_final_gan}
    \begin{aligned}
        \min_{\btheta\in\mathcal{I}^l}&V(\bm{D}_{\bphi^*},\Gtheta)=\mathbb{E}_{\x\sim\mathbb{P}_{H}}\left[\log(\mathbb{P}_l(\x))\right]\\
        &+\mathbb{E}_{\x\sim\mathbb{P}_{\Gtheta}}\left[\log(1-\mathbb{P}_l(\x))\right].
    \end{aligned} 
\end{equation}
\end{theorem}

Let $\mathbb{P}_{\overline{l}}=1-\mathbb{P}_l$ denote the probability that $l$ does not occur in sample $\x$. As $\Gtheta$ is not involved in  $\mathbb{E}_{\x\sim\mathbb{P}_{H}}\left[\log(\mathbb{P}_l(\x))\right]$, we only need to solve
\begin{equation}\label{eq:hypo_ce}
    \min_{\btheta\in\mathcal{I}^l}\mathbb{E}_{\x\sim\mathbb{P}_{\Gtheta}}\left[\log(\mathbb{P}_{\overline{l}}(\x))\right]=-H(\mathbb{P}_{\Gtheta},\mathbb{P}_{\overline{l}}),
\end{equation}
where $H$ is the cross entropy function \cite{shore1981properties,de2005tutorial,murphy2012machine}. Solving this problem means to maximize the information  of knowledge $l$ while keeping the generated distribution indistinguishable from the real data on the remaining parts. Intuitively, this procedure just points to the desired hypothetical distribution as is claimed by Theorem \ref{th:optD}.

\paragraph{Investigate Assumption \ref{as:ind} from the remaining knowledge.} Here the main difficulty is to handle the constraint $\btheta\in\mathcal{I}^l$ in Eq. (\ref{eq:hypo_ce}). To this end, we review the Indistinguishable Assumption \ref{as:ind} from the perspective of the remaining knowledge. An equivalent statement to ``$\mathbb{P}_{\Gtheta}$ and $\mathbb{P}_{H}$ are indistinguishable except for the knowledge of interest $l$'' is that, the remaining knowledge $r$ is not changed from $\mathbb{P}_{\Gtheta}$ to $\mathbb{P}_{H}$, and all the changes occur for the knowledge of interest $l$. Let $f^r_{\Gtheta}(\x), f^r_{H}(\x),f^r_{\mathcal{X}}(\x)$ be the remaining knowledge on sample $\x$ of distribution $\mathbb{P}_{\Gtheta},\mathbb{P}_H,\mathbb{P}_{\mathcal{X}}$, respectively. We should have an equivalent definition to the indiscernibility space
\begin{equation}\label{eq:eq_def_ind}
   \mathcal{I}^l=\{\btheta:\forall\x\in\mathcal{X}, f_{\Gtheta}^r(\x)=f^r_H(\x)=f_{\mathcal{X}}^r(\x)\}.
\end{equation}
\begin{remark}
We omit the discussion of the exact form of the remaining knowledge $f^r$ in order to bypass the demand of SCMs or strong ignorability that all confounders are known. Thus, our method does not rely on the exact factor factorization to the data distribution.
\end{remark}

Thus, the final objective can be transformed into
\begin{equation}\label{eq:hypo_final_obj}
     \min_{f_{\btheta}^r=f_{\mathcal{X}}^r}-H(\mathbb{P}_{\Gtheta},\mathbb{P}_{\overline{l}}).
\end{equation}
However, we still have the constraint $f_{\btheta}^r=f_{\mathcal{X}}^r$ unknown as we do not know the exact form of $f_{\btheta}^r$. Hopefully, the solution to this problem can be efficiently estimated. In the next section, we will study the associated numerical approximation.
\begin{remark}
If the constraint $\btheta\in\mathcal{I}^l$ is dropped, then the cross entropy achieves its optimal value if and only if $\mathbb{P}_{\Gtheta}=\mathbb{P}_{\overline{l}}$ \cite{murphy2012machine}. This is a degenerate case that the generator may even not yield the valid synthesis, and lose all knowledge except the one of interest $l$. Thus, enforcing the optimization inside the indiscernibility space is a decisive condition for knowledge extrapolation.
\end{remark}

\begin{algorithm}[t]
	\caption{Principal Knowledge Descent (PKD)}
	\label{algorithm:PKD}
	\begin{algorithmic}
		\STATE {\bfseries Input:} Maximum number of iteration $K$, pretrained generator $\bm{G}_{\btheta^{\mathcal{X}}}$ which captures the data distribution with $\mathbb{P}_{\bm{G}_{\btheta^{\mathcal{X}}}}=\mathbb{P}_{\mathcal{X}}$, prior distribution $\mathcal{N}(\mathcal{O},\mathcal{I})$ of the generator, posterior distribution $\mathbb{P}_l(\x)$ for knowledge $l$, step size $\epsilon$, batch size $m$, and hyper-parameter $\lambda>0$.
		\STATE {\bfseries Set:} $k=0$ and $\btheta^k=\btheta^{\mathcal{X}}$.
		\REPEAT
		        \STATE Randomly sample latent codes $\z_1,\dots,\z_m$ from prior distribution $\mathcal{N}(\mathcal{O},\mathcal{I})$.
				\STATE Compute $\vec{\bm{n}}=\nabla_{\btheta}\frac{1}{m}\sum_{i=1}^m\log(1-\mathbb{P}_l(\bm{G}_{\btheta^k}(\z_i));$
				\STATE Compute $\bm{I}=(\vert \vec{\bm{n}}\vert>\lambda)_b ,\bm{sgn}=(\vec{\bm{n}}>0)_b-(\vec{\bm{n}}<0)_b$, where $(\cdot)_b$ is element-wise Boolean operation;
		\STATE Update $\btheta^{k+1}=\btheta^k+\epsilon\bm{sgn}*\bm{I}, k=k+1$, where $*$ denotes element-wise multiplication;
		\UNTIL{ $k=K$.}
	\STATE {\bfseries Output:} An extrapolated generator $\bm{G}_{\btheta^K}$ that induces distribution $\mathbb{P}_{\bm{G}_{\btheta}^K}$ to estimate the hypothetical distribution $\mathbb{P}_H$.
	\end{algorithmic} 
\end{algorithm}

\subsection{Principal Knowledge Descent}
In this section, we study how to numerically solve Problem (\ref{eq:hypo_final_obj}). We show that its solution can be approximated through a series of principal knowledge descent with sparse and convex regularization. 

Given the current parameter $\btheta$, here we want to find a direction $\Delta\btheta$ such that
\begin{itemize}
    \item the knowledge of interest is altered accordingly, meaning that the cross entropy $-H(\mathbb{P}_{\bm{G}_{\btheta+\Delta\btheta}},\mathbb{P}_{\overline{l}})$ is optimized;
    \item the other knowledge is unchanged, \ieno, $\vert f_{\bm{G}_{\btheta+\Delta\btheta}}^r-f_{\mathcal{X}}^r\vert$  is as small as possible.
\end{itemize}
Such a direction can preserve $\btheta+\Delta\btheta$ to stay in $\mathcal{I}_{\mathcal{X}}^l$, and decline the value of the objective (\ref{eq:hypo_ce}) if $\btheta\in\mathcal{I}_{\mathcal{X}}^l$. We call this direction as the principal knowledge descent direction. With this method, we can compute a path starting from $\btheta^{\mathcal{X}}$, and along this path, the other knowledge is kept intact, but the cross entropy criterion increases drastically. Then any point at this path corresponds to a certain degree of altering the knowledge of interest $l$. Now we discuss how to compute the principal knowledge descent direction.

\paragraph{Remaining Knowledge Penalty.}Suppose $\btheta\in\mathcal{I}_{\mathcal{X}}^l$, then we have $f_{\Gtheta}^r=f_{\mathcal{X}}^r$ according to Eq. (\ref{eq:eq_def_ind}). Thus the change of the other knowledge under a small perturbation $\Delta\btheta$ can be written as
\begin{equation}\label{eq:PKD_motivation}
\begin{aligned}
        &\vert f_{\bm{G}_{\btheta+\Delta\btheta}}^r-f_{\mathcal{X}}^r\vert =\vert f_{\bm{G}_{\btheta+\Delta\btheta}}^r-f_{\Gtheta}^r\vert\\
         &\approx\vert(\nabla_{\btheta}f_{\Gtheta}^r)^T\Delta\btheta\vert
        \leq\Vert\nabla_{\btheta}f_{\Gtheta}^r\Vert_{\infty}\Vert\Delta\btheta\Vert_1\\
        &\qquad \qquad \qquad \qquad \leq L\Vert\Delta\btheta\Vert_1,
\end{aligned}
\end{equation}
provided that $\mathbb{P}_{\Gtheta}$ is continuously differentiable, and $L$ is an upper bound for $\Vert\nabla_{\btheta}f_{\Gtheta}^r\Vert_{\infty}$ on the indiscernibility space. The penult inequality stems from the Cauchy inequality, which offers an estimation to the upper bound of the other knowledge alteration. Thus, if we constrain the $\ell_1$ norm of $\Delta\btheta$, then we can constrain the overall change of the other knowledge. 

\paragraph{Linear Principal Part.} On the other hand, the descent value caused by updating $\btheta$ with $\btheta+\Delta\btheta$ is
\begin{equation}
\begin{aligned}
        &-H(\mathbb{P}_{\bm{G}_{\btheta+\Delta\btheta}},\mathbb{P}_{\overline{l}})+H(\mathbb{P}_{\bm{G}_{\btheta}},\mathbb{P}_{\overline{l}})\\
        =&-\nabla_{\btheta}H(\mathbb{P}_{\bm{G}_{\btheta_0}},\mathbb{P}_{\overline{l}})^T\Delta\btheta + o(\Vert\Delta\btheta\Vert_{\infty}).
\end{aligned}
\end{equation}
Thus, if we limit $\Vert\Delta\btheta\Vert_{\infty}\leq\epsilon\ll1$, then maximizing the descent value  $-H(\mathbb{P}_{\bm{G}_{\btheta+\Delta\btheta}},\mathbb{P}_{\overline{l}})+H(\mathbb{P}_{\bm{G}_{\btheta}},\mathbb{P}_{\overline{l}})$ is equivalent to minimizing the linear principal part $\nabla_{\btheta}H(\mathbb{P}_{\bm{G}_{\btheta_0}},\mathbb{P}_{\overline{l}})^T\Delta\btheta$.

\paragraph{One Step Objective.}In conclusion, we propose to get the descent direction of a single step from combining the remaining knowledge penalty and the linear principal part, \ieno
\begin{equation}\label{eq:PKD}
\begin{aligned}
        \min_{-\epsilon\preceq \Delta\btheta\preceq \epsilon}\nabla_{\btheta}H(\mathbb{P}_{\bm{G}_{\btheta_0}},\mathbb{P}_{\overline{l}})^T\Delta\btheta
        +\lambda\Vert\Delta\btheta\Vert_1,
\end{aligned}
\end{equation}
where $0<\epsilon\ll1$ is the step size, $\lambda=L\lambda_0>0$ containing two factors of an estimation $L$ to the upper bound of $\Vert\nabla_{\btheta}f_{\bm{G}_{\btheta_0}}^r\Vert_{\infty}$ and a hyper-parameter $\lambda_0$ to adjust the weight of the regularization term $\Vert\Delta\btheta\Vert_1$, and  $-\epsilon\preceq\Delta\btheta\preceq\epsilon$ means that each entry of $\Delta\btheta$ lies in  $[-\epsilon,\epsilon]$.
%
The first term $\nabla_{\btheta}H(\mathbb{P}_{\bm{G}_{\btheta_0}},\mathbb{P}_{\overline{l}})^T\Delta\btheta$ in (\ref{eq:PKD}) maximizes the descent value $(H(\mathbb{P}_{\bm{G}_{\btheta+\Delta\btheta}},\mathbb{P}_{\overline{l}})-H(\mathbb{P}_{\bm{G}_{\btheta}},\mathbb{P}_{\overline{l}}))$ induced by $\Delta\btheta$,  moving $\mathbb{P}_{\Gtheta}$ closer to $\mathbb{P}_{\overline{l}}$ as it is the cross entropy term. The second term $\lambda\Vert\Delta\btheta\Vert_1$ penalizes the overall change $\Vert\nabla_{\btheta}f_{\Gtheta}^r\Vert_{\infty}\Vert\Delta\btheta\Vert_1$ induced to the other knowledge. 

\paragraph{Sparsity of the Solution.}
The $\ell_1$ penalty of $\Vert\Delta\btheta\Vert_1$ is well known to enforce sparsity to the elements of the solution \cite{santosa1986linear, tibshirani1996regression}. Namely, only a small number of elements of $\Delta\btheta$ will be non-zeros, meaning that the principal knowledge descent will only involve a fraction of parameters, while the most parameters of the generator will remain unchanged. The typical knowledge, like the expression, age, or gender of facial images, may only contain information of low dimensions \cite{penev2000global, harkonen2020ganspace, shen2020interfacegan}. It will be obviously excessive for extracting a single knowledge domain in the whole parameter space. Accompanying with the excessively used parameters are two well-known challenges: over-fitting \cite{anderson2004model} and numerical instability \cite{hildebrand1987introduction}. The expressive capability of millions of parameters of the generator is powerful, thereby easily overfitting the bias of the classification or regression network we use as the optimal discriminator. Another issue is the numerical instability. As we have mentioned, the valid parameters relevant to a certain semantic feature should be sparse due to low-dimensional semantic information. Thus, updating all the parameters simultaneously may bring unexpected changes like entanglement of attributes. For example, changing `age'  leads to adding `beard' in facial images.

\paragraph{Closed-Form Solution.} An intriguing property pertaining to Problem (\ref{eq:PKD}) is that this convex optimization problem has a closed-form solution by virtue of strong duality and Karush–Kuhn–Tucker conditions \cite{boyd2004convex}.
\begin{theorem}\label{th:solution}
There is $\lambda_{max}>0$ such that $\forall\lambda\in(0,\lambda_{max})$, Problem (\ref{eq:PKD}) admits a non-empty solution set. Specifically,  a special solution can be attained by
\begin{equation}\label{eq:screening}
    \left\{\begin{array}{ll}
         \Delta\btheta_i=0,& \text{if}~\vert\nabla_{\btheta}H_i\vert\leq\lambda, \\
         \Delta\btheta_i=\epsilon,&\text{if}~\nabla_{\btheta}H_i<0, 0\leq\lambda<\vert\nabla_{\btheta}H_i\vert,\\
         \Delta\btheta_i=-\epsilon,&\text{if}~\nabla_{\btheta}H_i>0, 0\leq\lambda<\vert\nabla_{\btheta}H_i\vert,
    \end{array}\right.
\end{equation}
where $\Delta\btheta_i$ and $\nabla_{\btheta}H_i=[\nabla_{\btheta}H(\mathbb{P}_{\bm{G}_{\btheta_0}},\mathbb{P}_{\overline{l}})]_i$ are the $i$-th element of $\Delta\btheta$ and $\nabla_{\btheta}H(\mathbb{P}_{\bm{G}_{\btheta_0}},\mathbb{P}_{\overline{l}})$, respectively.
\end{theorem}
\begin{figure*}
    \centering
    \includegraphics[width=1.0\linewidth]{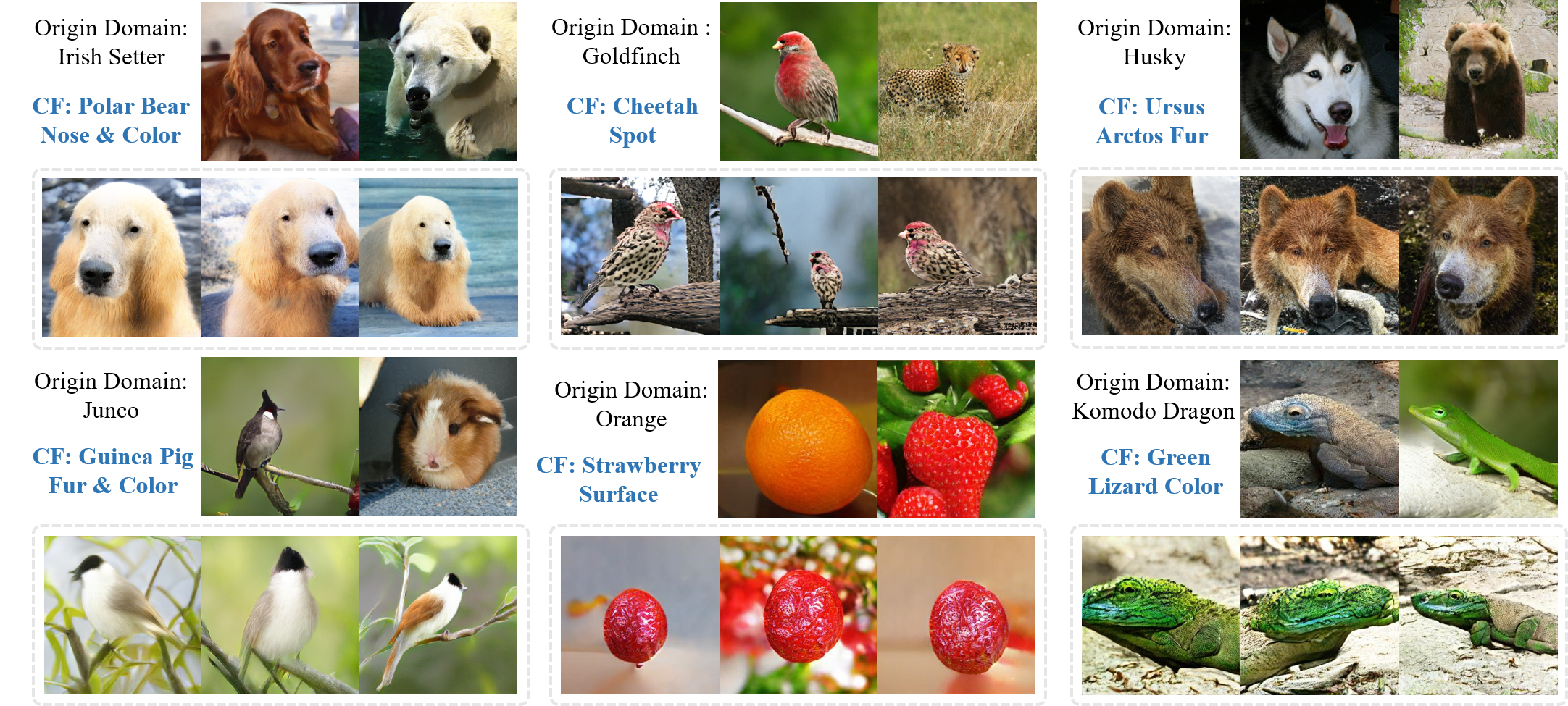}\vspace{-0.4cm}
    \caption{Knowledge extrapolation of BigGAN on ImageNet. Odd rows report the original data domain and extrapolated counterfactual knowledge (marked with \textcolor{cf}{\textbf{CF:}} and \textcolor{cf}{\textbf{purple}} color), while even rows report the counterfactual results generated by the proposed method. To the best of our knowledge, this is the first work that yields high-fidelity photo-realistic counterfactual synthesis of various image domains.}
    \label{fig:cf_biggan}\vspace{-0.2cm}
\end{figure*}

\paragraph{Principal Knowledge Descent.} We conclude our algorithm to solve Problem (\ref{eq:hypo_ce}) in Algorithm \ref{algorithm:PKD}. This algorithm manages to minimize the objective of Problem (\ref{eq:hypo_ce}) while trying to maintain the remaining knowledge distribution. Rigorously, we have the following theorem.
\begin{theorem}\label{th:lambda}
Let $\Delta$ be the descent value of the objective (\ref{eq:hypo_ce}) by implementing Algorithm \ref{algorithm:PKD}, and $\delta$ be the change of the other knowledge, \ieno,
\begin{equation}
    \Delta=H(\mathbb{P}_{\bm{G}_{\btheta^K}},\mathbb{P}_{\overline{l}})-H(\mathbb{P}_{\bm{G}_{\mathcal{X}}},\mathbb{P}_{\overline{l}}),\vspace{-0.2cm}
\end{equation}
\begin{equation}
    \delta=\mathbb{E}_{\x\sim\mathbb{P}_{\mathcal{X}}}\left[\left\vert f_{\bm{G}_{\btheta^K}}^r(\x)-f_{\mathcal{X}}^r(\x)\right\vert\right],
\end{equation}
where $K$ is the iteration turns. 
Assume that $\mathbb{P}_{\bm{G}_{\btheta^{\mathcal{X}}}}=\mathbb{P}_{\mathcal{X}}$, $\epsilon$ is small enough, and $L=\sup_{\Vert\btheta-\btheta^{\mathcal{X}}\Vert_{\infty}\leq K\epsilon}\Vert\nabla_{\btheta}f_{\Gtheta}^r\Vert_{\infty}$. There is  $\lambda_{max}>0$ such that $\forall\lambda\in(0,\lambda_{max})$, we have $\Delta>0$ and
\begin{equation}
    \frac{\Delta}{\delta}\geq \frac{\lambda}{L}+o(1).
\end{equation}
\end{theorem}
This theorem tells us that choosing a large $\lambda$ can yield small variation of the remaining knowledge. Recall that $\lambda=\lambda_0L$. This theorem also implies that  $\lambda_0$ controls the principal ratio of Algorithm \ref{algorithm:PKD} --- the ratio of change in the knowledge of interest and the other knowledge.

\subsection{Dirac Knowledge Extrapolation}
While our method is designed for distribution-wise extrapolation, we can also use it for single image extrapolation by setting the data distribution as the Dirac distribution \cite{arfken1999mathematical}
\begin{equation}
   \mathbb{P}_{\mathcal{X}}(\x)=\delta_{\x_0}(\x)=\left\{\begin{array}{ll}
       1 &\x=\x_0,  \\
        0 &\x\neq\x_0.
    \end{array}\right.
\end{equation}
We may then change the prior distribution $\mathcal{N}(\mathcal{O},\mathcal{I})$ in Algorithm \ref{algorithm:PKD} with $\delta_{\GthetaX^{-1}(\x_0)}(\z)$ to implement the single image knowledge extrapolation. While the Dirac distribution is not smooth, directly implementing it may cause numerical instability. To enhance numerical stability, we will instead change the prior distribution $\mathcal{N}(\mathcal{O},\mathcal{I})$ in Algorithm \ref{algorithm:PKD} with $\mathcal{N}(\GthetaX^{-1}(\x)+\mathcal{O},\xi\mathcal{I})$, where $0<\xi\ll1$ is a small number to approximate the Dirac distribution.

\begin{figure*}
    \centering
    \includegraphics[width=1.0\linewidth]{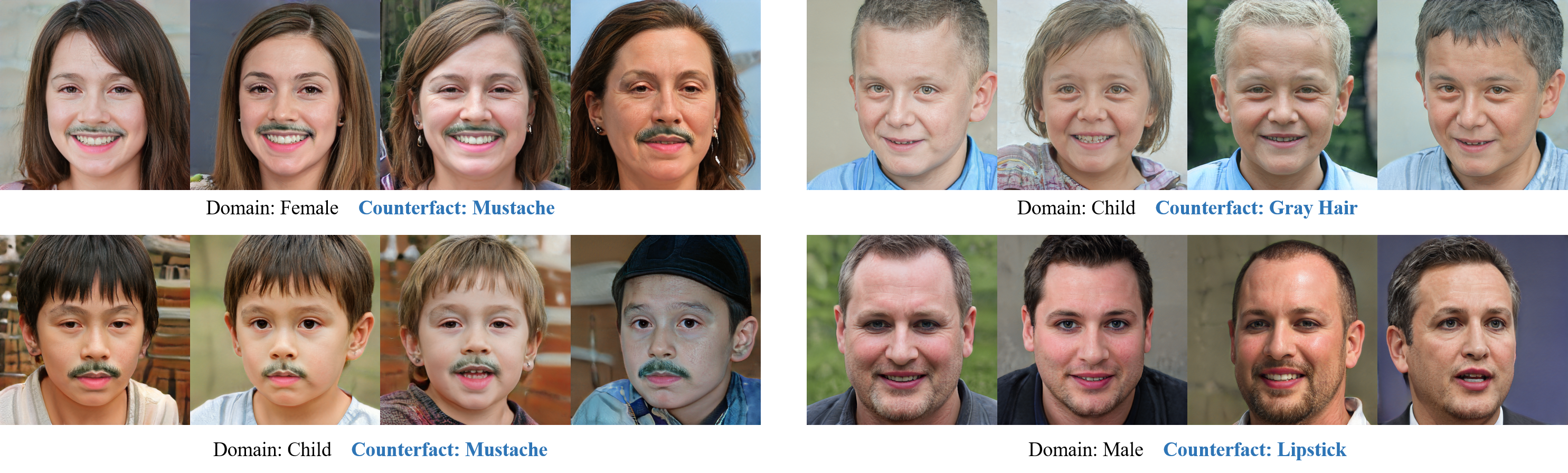}\vspace{-0.4cm}
    \caption{Knowledge extrapolation of StyleGAN2 on FFHQ dataset. Odd rows report the original data domain and extrapolated counterfactual knowledge (marked in \textcolor{cf}{\textbf{purple}}), while even rows report the counterfactual results generated by the proposed method.}
    \label{fig:ffhq_cf}\vspace{-0.4cm}
\end{figure*}
\begin{figure}
    \centering
    \includegraphics[width=1.0\linewidth]{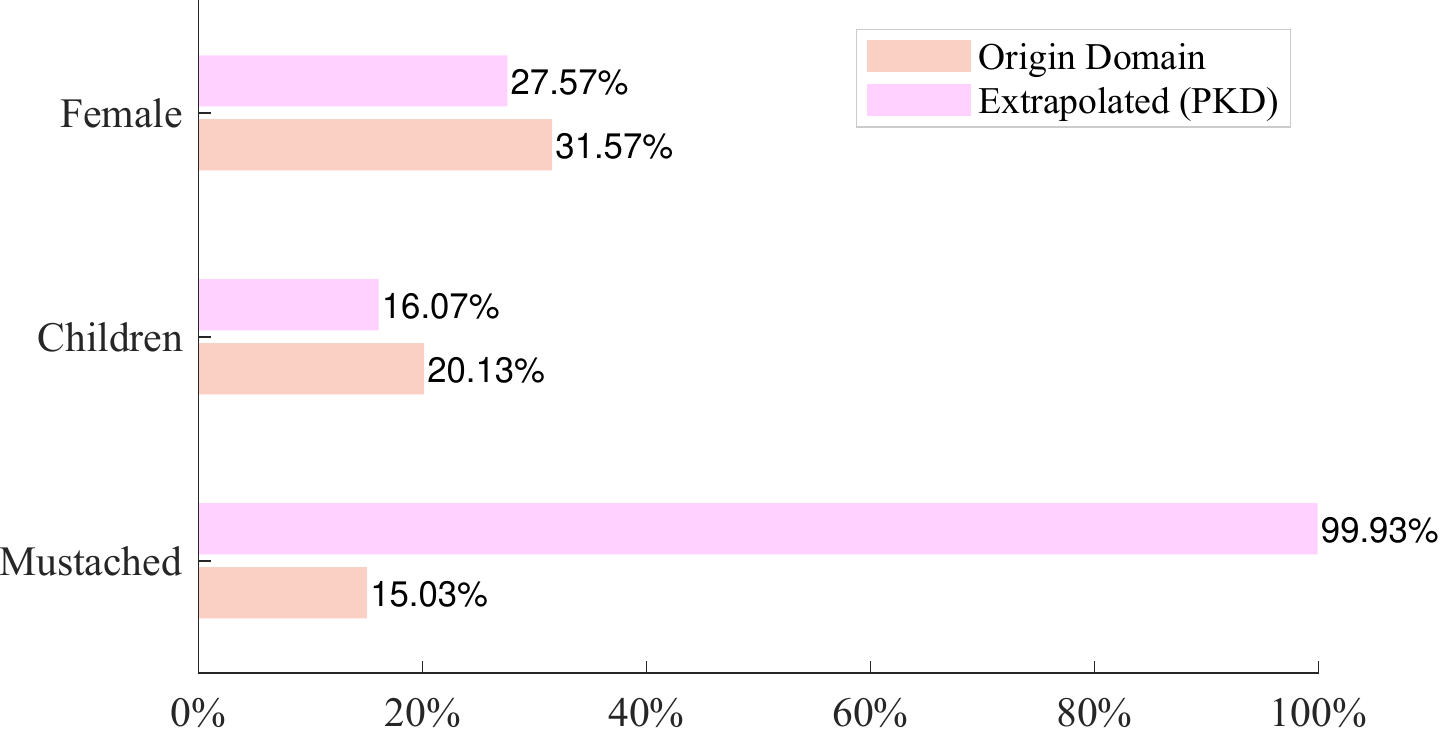}\vspace{-0.4cm}
    \caption{Statistics of the original distribution and extrapolated distribution after extrapolating `mustache' on FFHQ faces. PKD increases the ratio of faces with `mustache' from 15.03\% to 99.93\%, while only inducing less than 5\% faces becoming `male' or 'mature'. Considering the subtle case where `mustache' alone can confuse age or gender, the consequence of entanglement is trivial. Thus, almost all faces in the extrapolated distribution are mustached, while the other statistics like the ratio of children and female are well preserved.}
    \label{fig:user_study}\vspace{-0.4cm}
\end{figure}
\begin{table}[t]
    \centering
    \begin{tabular}{cccc}
    \toprule
        Metric & FID&  IS & Path Length \\
        \midrule
        Original GAN & \textbf{5.31}& \textbf{4.36} & 185.59\\
        PKD-Mustache & 7.17 & 4.10& 187.50\\
        PKD-Lipstick & 7.26 & 4.26& \textbf{175.20}\\
        PKD-GrayHair & 7.04 & 4.18 & 198.24\\
        \bottomrule
    \end{tabular}\vspace{-0.2cm}
    \caption{Numerical metrics for the quality of extrapolated distributions of PKD method on the FFHQ domain. Basically, we find that the decline of synthesis quality is negligible.}
    \label{tab:is_ppl}\vspace{-0.4cm}
\end{table}

\section{Findings and Results}
In this section, we present several discoveries from our proposed Principal Knowledge Descent (PKD) method and extrapolation results of state-of-the-art GANs, including BigGAN256-Deep \cite{brock2018large}, StyleGAN2 \cite{karras2019style,karras2020analyzing} on FFHQ faces \cite{karras2019style}, and StyleGAN2-ADA \cite{karras2020training} on  BreCaHAD \cite{aksac2019brecahad} which contains breath cancer slices. The details of the experiments are reported in Appendix Sec. \ref{sec:exp_set}, including choices of the hyper-parameters $\lambda,\epsilon, K, m$, sources of all pretrained models (\ieno, generators and posterior estimation models), and dataset information. We also investigate the influence of $\lambda$ to the extrapolation and model sparsity in Appendix Sec. \ref{sec:lambda}.

\paragraph{PKD offers a new methodology for counterfactual synthesis.}
We report knowledge extrapolation results of our PKD method on ImageNet \cite{deng2009imagenet} data domain and FFHQ \cite{karras2019style} face data domain in Fig. \ref{fig:cf_biggan} and \ref{fig:ffhq_cf}, respectively. In ImageNet domain, we use a pretrained BigGAN256-Deep model as the pretrained generator and ResNet50 \cite{he2016deep} classifiers as the posterior distribution for knowledge of interest. We infer the results of counterfactual combination of knowledge among different ImageNet categories. As displayed in Fig. \ref{fig:cf_biggan}, we successfully synthesize non-existent species such as goldfinches with cheetah spot, huskies with ursus arctos fur, oranges with strawberry surface, \etcno. In FFHQ facial images domain, we use the StyleGAN2 model as the pretrained generator, and ResNet50 classifiers trained on CelebA-HQ \cite{karras2018progressive} annotations for facial attributes like `mustache', `lipstick', `gray hair' as the posterior distribution for knowledge of interest. We infer the results of altering distributions of those facial attributes in Fig. \ref{fig:ffhq_cf}, which shows counterfactual images such as women and children in `mustache', men in `lipstick', and children in `gray hair'. As neural networks are disable to detect counterfactual results, we further conduct user study to confirm that whether the knowledge extrapolation globally succeeds. To do this, we randomly sample 3,000 images from the original generator distribution and the extrapolated distribution after extrapolating `mustache' on FFHQ faces, respectively. We mix and randomly shuffle those images, and then ask testers three questions for each image: whether the person is 1) a child, 2) a female, and 3) mustached (Refer to Appendix for details of the user study). The result is reported in Fig. \ref{fig:user_study}. We can find that PKD only significantly changes knowledge `mustache' while maintaining the two \textit{confounders} `child' and `female' nearly intact. Overall, the proposed PKD method can generate high-fidelity counterfactual results, although the exact causality relations are not pre-defined. Thus, counterfactual synthesis may not rely on the prior SCM to guide the generator. We demonstrate that PKD is also a competitive alternative for this task.

\begin{figure}
    \centering
    \includegraphics[width=1.0\linewidth]{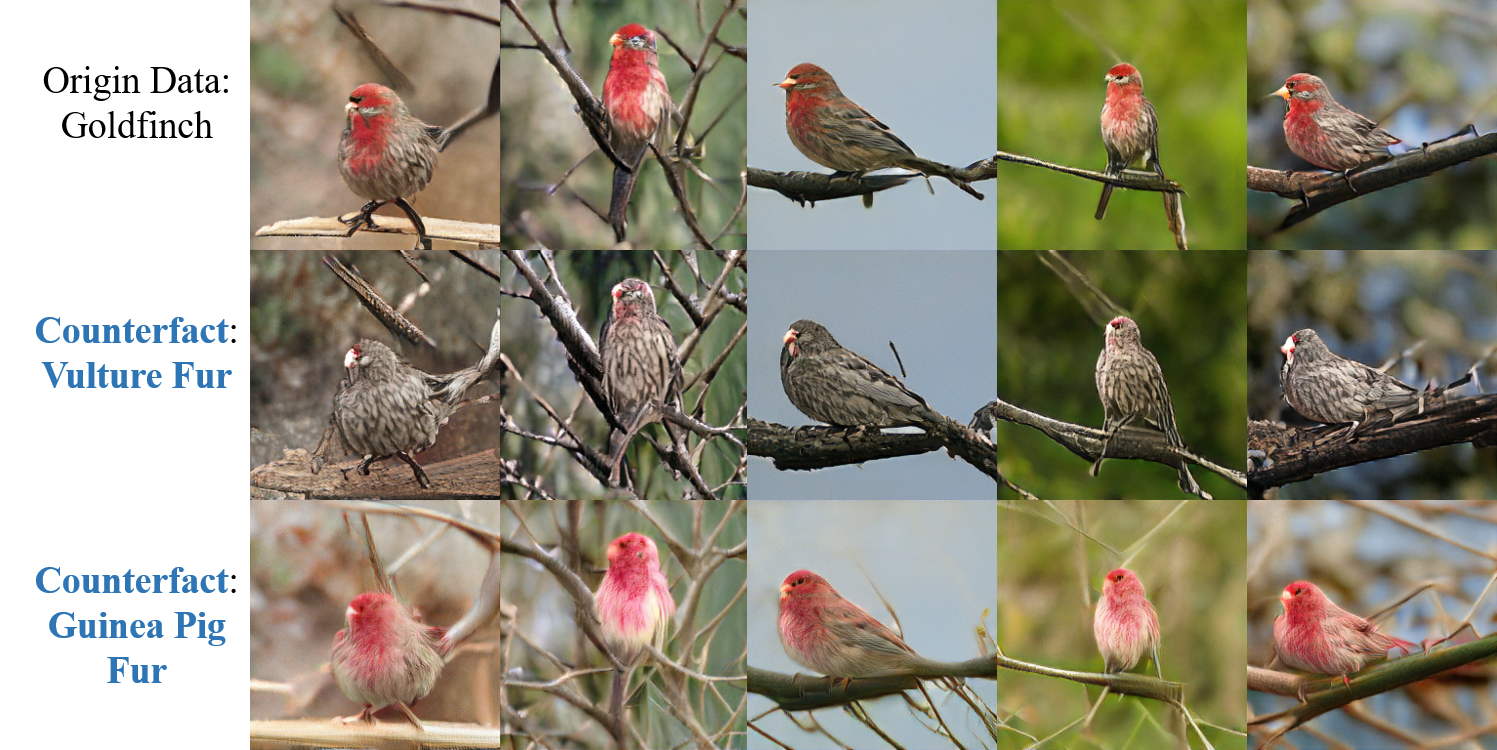}\vspace{-0.4cm}
    \caption{Dirac knowledge extrapolation to ImageNet animals. Top row shows the original data, middle and bottom rows show the extrapolated data, with extrapolated knowledge marked in \textcolor{cf}{\textbf{purple}}.}
    \label{fig:biggan_dirac}\vspace{-0.4cm}
\end{figure}
\begin{figure*}
    \centering
    \includegraphics[width=1.0\linewidth]{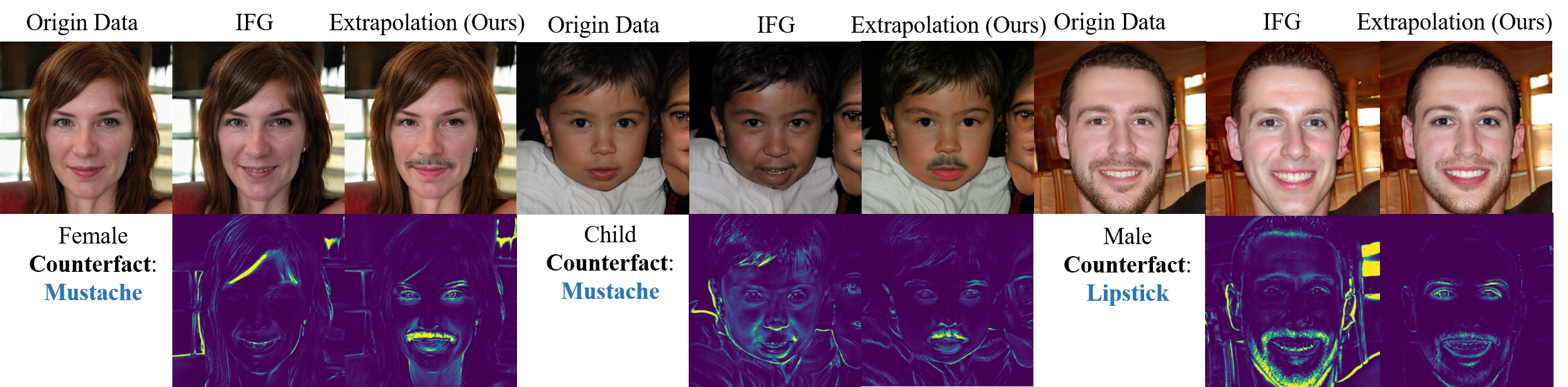}\vspace{-0.4cm}
    \caption{Counterfactual synthesis of latent image editing method (IFG) and our Dirac knowledge extrapolation method.  Variations of image pixels are highlighted in the bottom row. Latent image editing method fails the counterfactual inference of women and children in mustache, removing mustache when extrapolating lipstick, while our method can easily handle these cases and induce changes more concentrated in the regions of interests.}\vspace{-0.4cm}
    \label{fig:ifgvpkd}
\end{figure*}

\definecolor{light green}{RGB}{240, 251, 185}
\begin{figure*}
    \centering
    \includegraphics[width=1.0\linewidth]{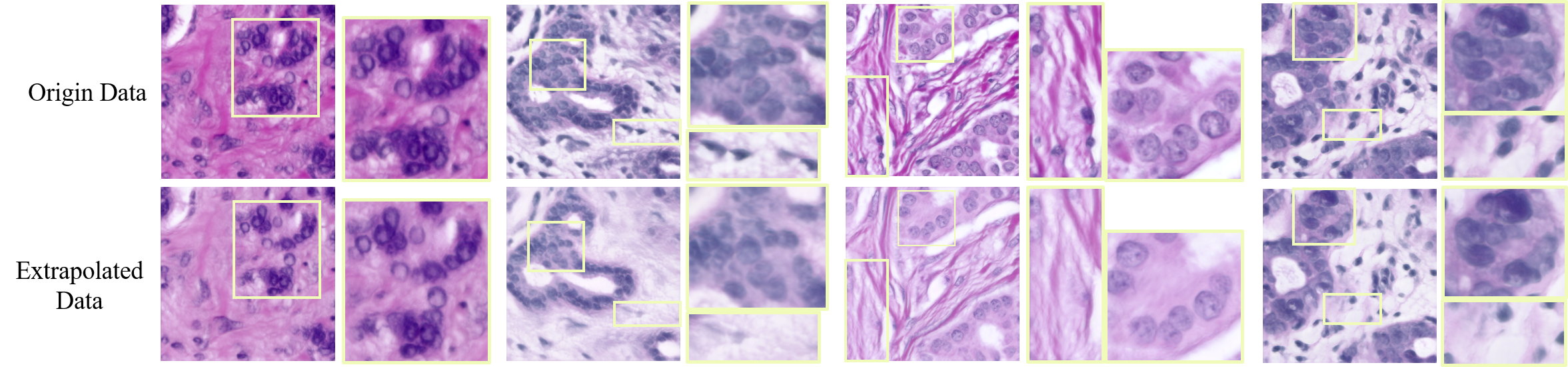}\vspace{-0.4cm}
    \caption{Few-shot Dirac knowledge extrapolation of tissue slices of breath cancer patients. The top row shows the original slices. The tissue slices are stained by H\&E \cite{dapson2009dyes}, containing bluish violet multi-core structures of the cancer nucleus and light red color of extracellular materials. Here we extrapolate each slice toward milder symptoms of cancer. The results are reported in the bottom row. In \textit{most local regions} of the slices, the number of cancer nucleus decreases significantly, details are zoomed in with \colorbox{pink}{\textcolor{light green}{light green}} boxes. The whole training data to support our method is composed of merely \textit{162} annotated slice images.}
    \label{fig:extra_tumor}\vspace{-0.4cm}
\end{figure*}

\paragraph{Dirac Knowledge Extrapolation is very powerful.} We report the results of Dirac Knowledge Extrapolation of PKD in Fig. \ref{fig:biggan_dirac}, \ref{fig:ifgvpkd}, and \ref{fig:extra_tumor}. The Dirac Knowledge Extrapolation focuses on extrapolating knowledge of a given image rather than the whole distribution. For ImageNet and FFHQ data domains, the pretrained models and posterior distributions are selected as in previous knowledge extrapolation, and the results are reported in Fig. \ref{fig:biggan_dirac} and \ref{fig:ifgvpkd}, respectively. Here we are interested in comparing the knowledge extrapolation with the recent latent image editing methods \cite{shen2020interfacegan,patashnik2021styleclip,tewari2020pie}, which edit facial attributes by attribute vectors in latent spaces of pretrained GANs. Despite the success in editing usual attributes, those methods are generally weak when conducting counterfactual synthesis. A major reason is that their synthesis will always lie in the pretrained generator distribution, as the parameters of the GAN model are not selected. The pretrained generator distribution can hardly yield counterfactual results. Specifically, we report the comparison with a baseline latent image editing method called InterFaceGAN (IFG) \cite{shen2020interfacegan} in Fig. \ref{fig:ifgvpkd} (More comparisons are given in Appendix). We further conduct Dirac knowledge extrapolation in the BreCaHAD data domain. This dataset consists of 162 slice images and each of them has annotations for cancer nuclear. We train a ResNet50 classifier to infer the posterior probability for cancer severity of a tissue slice image on the few-shot annotation images, and use the StyleGAN2-ADA model trained on this dataset as the pretrained generator. We infer the results of reducing cancer severity in Fig. \ref{fig:extra_tumor}. Despite the few-shot training data, the PKD method can successfully infer the expected results. In all cases, the Dirac Knowledge extrapolation demonstrates compelling performance.

\paragraph{PKD is efficient and flexible when applied to SOTA models.} The fidelity of counterfactual synthesis of PKD significantly surpasses previous causal GAN works. The major reason is its flexibility to combine with state-of-the-art generative models. Previous methods need to adapt generator architectures to be compatible with their prior SCMs or causal graphs, which is a non-trivial task. To the best of our knowledge, this is the first work that is capable of conducting counterfactual synthesis of photo-realistic effect without changing the generator architecture. Moreover, we find that the PKD method is very efficient when applying to SOTA models, \ieno, StyleGAN2 and BigGAN. We find that appealing counterfactual synthesis in most cases can be attained within 10-20 PKD iterations, and the damage to the synthesis fidelity caused by PKD is negligible. In Tab. \ref{tab:is_ppl}, we report numerical metrics (\ieno, Fr\'echet Inception Distance (FID), Inception Score (IS), and Path Length as in \cite{karras2019style}) of synthesis quality after knowledge extrapolation, indicating that the influence to synthesis quality is little.

\section{Conclusion}
This paper studies the problem of knowledge extrapolation of GANs, where the original generated distribution of a pretrained GAN is altered under the guidance of a novel Principal Knowledge Descent method to obtain counterfactual synthesis. Different from traditional methods that conduct counterfactual synthesis based on prior SCMs, this paper proposes to leverage a simple assumption that the extrapolated distribution and the original distribution are indistinguishable except for the knowledge of interest. Thus, our work gets rid of the usual demands of traditional methods to change the generator architecture to obey prior causalities. As a result, the proposed method is much more convenient to apply to SOTA generator models, and can yield much more photo-realistic counterfactual results.


\bibliography{example_paper}
\bibliographystyle{icml2022}

\newpage
\appendix
\onecolumn
\section{Proof to Theorems}
\subsection{Theorem \ref{th:optD}}
\begin{proof}
It is well known that the optimal discriminator is \cite{goodfellow2014generative}
\begin{equation}
    \bm{D}_{\bphi^*}=\frac{\mathbb{P}_H}{\mathbb{P}_H+\mathbb{P}_{\Gtheta}}.
\end{equation}
When $\btheta\in\mathcal{I}^l$, we have some probability distribution $\mathbb{P}_l$ for knowledge $l$ such that
\begin{equation}
    \mathbb{P}_{\Gtheta}=\frac{1-\mathbb{P}_l}{\mathbb{P}_l}\mathbb{P}_{H}.
\end{equation}
Thus we have
\begin{equation}
    \bm{D}_{\bphi^*}=\frac{\mathbb{P}_H}{\mathbb{P}_H+\frac{1-\mathbb{P}_l}{\mathbb{P}_l}\mathbb{P}_{H}}=\mathbb{P}_l.
\end{equation}
\end{proof}

\subsection{Theorem \ref{th:solution}}
\begin{proof}
\newcommand{\bbeta}{\bm{\beta}}
\newcommand{\bgamma}{\bm{\gamma}}
\newcommand{\V}{\bm{V}}
\newcommand{\n}{\bm{n}}

Let $\x=\Delta\btheta$, $\V=\nabla_{\btheta}H(\mathbb{P}_{\bm{G}_{\btheta_0}},\mathbb{P}_{\overline{l}})$, $N$ be the volume of parameters of the generator, $\bbeta=(\bbeta_1,...,\bbeta_N)^T$, $\bgamma=(\bgamma_1,...,\bgamma_N)^T$ be the Lagrangian multiplier vectors, and $\bm{1}=(1,...,1)^T$ be the vector of all ones. We write the Lagrangian dual function of Problem (\ref{eq:PKD})
\begin{gather}
    g(\bbeta,\bgamma)=\inf_{\x}L(\x,\bbeta,\bgamma)=\inf_{\x}\V^T\x+\lambda\Vert\x\Vert_1+\sum_{i=1}^N\bbeta_i(\x_i-\epsilon)+\sum_{i=1}^N\bgamma_i(-\epsilon-\x_i)\\
    =\inf_{\x}(\V+\bbeta-\bgamma)^T\x+\lambda\Vert\x\Vert_1-\epsilon(\bbeta+\bgamma)^T\bm{1},\\
    \text{s.t.} ~\bbeta_i\geq0,~\bgamma_i\geq0, ~i=1,\dots,N.
\end{gather}

As $L(\x,\bbeta,\bgamma)$ is a convex function, its optimal value $\x^*$ is reached if and only if
\begin{equation}
    0\in\partial_{\x} L(\x^*,\bbeta,\bgamma)=\V+\bbeta-\bgamma + \lambda\partial_{\x}\Vert\x^*\Vert_1,
\end{equation}
where $\partial_{\x}$ denotes the sub-gradient operator of convex function with respect to $\x$. Considering
\begin{equation}
    \partial_{\x}\Vert\x^*\Vert_1= \{ \n\in\mathbb{R}^N:\Vert\n\Vert_{\infty}\leq1 \},
\end{equation}
we have
\begin{equation}
    0\in \{\V+\bbeta-\bgamma+\lambda\n:\Vert \n\Vert_{\infty}\leq1\}.
\end{equation}
Thus we get that
\begin{equation}
    L(\x^*,\bbeta,\bgamma)=\left\{\begin{array}{ll}
         -\epsilon(\bbeta+\bgamma)^T\bm{1},&\text{if  }~\Vert\V+\bbeta-\bgamma\Vert_{\infty}\leq\lambda,  \\
        -\infty, & \text{otherwise.}
    \end{array} \right.
\end{equation}
Then, the Lagrangian duality form of Problem (\ref{eq:PKD}) is
\begin{gather}
    \min_{\bbeta,\bgamma}-\epsilon(\bbeta+\bgamma)^T\bm{1},\\
    \text{s.t.} ~ \bbeta_i\geq0,~\bgamma_i\geq0,~\vert\V_i+\bbeta_i-\bgamma_i\vert\leq\lambda.
\end{gather}
This problem can be easily solved by linear programming technique. A special solution $(\bbeta^*,\bgamma^*)$ can be obtained by
\begin{equation}\label{eq:appendix_dual}
    \left\{\begin{array}{ll}
        \bbeta^*_i=\bgamma^*_i=0, &\text{if} ~ -\lambda<\V_i<\lambda, \\
         \bbeta^*_i=0,\bgamma^*_i=\V_i-\lambda,&\text{if} ~ \V_i\geq\lambda,\\
         \bbeta^*_i=-\lambda-\V_i,\bgamma^*_i=0, &\text{if} ~ \V_i\leq-\lambda,
    \end{array}\right.
\end{equation}
for $i=1,\dots,N$. Recalling the Karush–Kuhn–Tucker conditions of convex optimization, we have that $\x^*,\bbeta^*,\bgamma^*$ satisfy the following conditions 
\begin{gather}
    \bbeta^*_i(\x^*_i-\epsilon)=0,~\bgamma^*_i(-\epsilon-\x^*_i)=0,~i=1,...,N,\\
    \vert\V_i+\bbeta^*_i-\bgamma^*_i\vert<\lambda\Rightarrow \x^*_i=0.
\end{gather}
We then have
\begin{equation}
    \left\{\begin{array}{ll}
         \x^*_i-\epsilon=0,&\text{if}~\bbeta^*_i\neq0, \\
         -\epsilon-\x^*_i=0,&\text{if}~\bgamma^*_i\neq0,\\ 
         \x^*_i=0,&\text{if}~\bbeta^*_i=\bgamma^*_i=0,~\vert\V_i\vert<\lambda.
    \end{array}\right.
\end{equation}
Combining Eq. (\ref{eq:appendix_dual}), we then conclude the theorem
\begin{equation}
    \left\{\begin{array}{ll}
         \x_i=0,&\text{if}~\vert\V_i\vert\leq\lambda, \\
         \x_i=\epsilon,&\text{if} ~\V_i<0, 0\leq\lambda<\vert\V_i\vert,\\
         \x_i=-\epsilon,&\text{if}~\V_i>0, 0\leq\lambda<\vert\V_i\vert.
    \end{array}\right.
\end{equation}
\end{proof}

\subsection{Theorem \ref{th:lambda}}
\begin{proof}
We first consider one step of Principal Knowledge Descent (PKD). Assume that the current step is $k$ and $k<K$. Then we define
\begin{equation}
    \Delta_k= H(\mathbb{P}_{\bm{G}_{\btheta^{k+1}}},\mathbb{P}_{\overline{l}})-H(\mathbb{P}_{\bm{G}_{\btheta^k}},\mathbb{P}_{\overline{l}})=\nabla_{\btheta}H(\mathbb{P}_{\bm{G}_{\btheta^k}},\mathbb{P}_{\overline{l}})^T\Delta\btheta^k+o(\epsilon),
\end{equation}
\begin{equation}
    \delta_k=\mathbb{E}_{\x\sim\mathbb{P}_{\mathcal{X}}}\left[\left\vert f^r_{\bm{G}_{\btheta^{k+1}}}-f^r_{\bm{G}_{\btheta^{k}}}\right\vert\right],
\end{equation}
where $\Delta\btheta^k$ is given as in Theorem \ref{th:solution}. 

We first prove that $\Delta_k>0$ for small $\epsilon$. This is obvious since among the non-zero elements of $\Delta\btheta$, we have $\Delta\btheta^k_i=\epsilon$ if $\nabla_{\btheta}H(\mathbb{P}_{\bm{G}_{\btheta^k}},\mathbb{P}_{\overline{l}})_i\Delta\btheta^k_i>0$, and $\Delta\btheta^k_i=-\epsilon$ if $\nabla_{\btheta}H(\mathbb{P}_{\bm{G}_{\btheta^k}},\mathbb{P}_{\overline{l}})_i\Delta\btheta^k_i<0$. Thus the term $\nabla_{\btheta}H(\mathbb{P}_{\bm{G}_{\btheta^k}},\mathbb{P}_{\overline{l}})^T\Delta\btheta^k$ must be positive. In fact, we have 
\begin{equation}
    \nabla_{\btheta}H(\mathbb{P}_{\bm{G}_{\btheta^k}},\mathbb{P}_{\overline{l}})^T\Delta\btheta^k\geq\lambda\epsilon,
\end{equation}
as long as $\Vert \nabla_{\btheta}H(\mathbb{P}_{\bm{G}_{\btheta^k}},\mathbb{P}_{\overline{l}})\Vert_{\infty}>\lambda$. While $o(\epsilon)$ is the high-order infinitesimal of $\epsilon$, we have 
\begin{equation}
    \Delta_k=\nabla_{\btheta}H(\mathbb{P}_{\bm{G}_{\btheta^k}},\mathbb{P}_{\overline{l}})^T\Delta\btheta^k+o(\epsilon)\geq\lambda\epsilon+o(\epsilon)>0.
\end{equation}

Assume that there are $M$ elements of $\Delta\btheta^k$ that are non-zero. Then we have
\begin{equation}
    \Delta_k=\sum_{i=1}^N \vert\nabla_{\btheta}H(\mathbb{P}_{\bm{G}_{\btheta^k}},\mathbb{P}_{\overline{l}})_i\Delta\btheta^k_i\vert+o(\epsilon)\geq M\lambda\epsilon+o(\epsilon),
\end{equation}
as the non-zero elements of $\Delta\btheta^k$ corresponds to elements of $\nabla_{\btheta}H(\mathbb{P}_{\bm{G}_{\btheta^k}},\mathbb{P}_{\overline{l}})$ that have absolute values larger than $\lambda$.

Note that
\begin{equation}
    \Vert\btheta^k-\btheta^{\mathcal{X}}\Vert_{\infty}\leq \sum_{i=1}^k\Vert\btheta^i-\btheta^{i-1}\Vert_{\infty}\leq k\epsilon< K\epsilon,
\end{equation}
where $\btheta^0=\btheta^{\mathcal{X}}$. Thus we have
\begin{equation}
    \sup_{\Vert \btheta-\btheta^k\Vert_{\infty}<\epsilon}\Vert\nabla_{\btheta}f^r_{\Gtheta}\Vert_{\infty}\leq L,
\end{equation}
as 
\begin{equation}
    \{\Vert \btheta-\btheta^k\Vert_{\infty}<\epsilon\}\in \{\Vert \btheta-\btheta^{\mathcal{X}}\Vert_{\infty}\leq K\epsilon\}.
\end{equation}
Thus we also have
\begin{equation}
    \delta_k\leq L\mathbb{E}_{\x\sim\mathbb{P}_{\mathcal{X}}}[\Vert\Delta\btheta\Vert_1]= L\mathbb{E}_{\x\sim\mathbb{P}_{\mathcal{X}}}[M\epsilon]=ML\epsilon. 
\end{equation}
Then we can conclude
\begin{equation}
    \frac{\Delta_k}{\delta_k}\geq\frac{M\lambda\epsilon}{ML\epsilon}+o(1)=\frac{\lambda}{L}+o(1).
\end{equation}
Note that 
\begin{equation}
    \delta\leq\delta_1+...+\delta_K,
\end{equation}
and 
\begin{equation}
    \Delta=\Delta_1+...+\Delta_K.
\end{equation}

We then have
\begin{gather}
    \frac{\Delta}{\delta}=\frac{\sum_{i=1}^K\Delta_K}{\delta}\geq\frac{\sum_{i=1}^K\Delta_i}{\sum_{i=1}^K\delta_i}\geq\frac{\sum_{i=1}^K (\frac{\lambda}{L}+o(1))\delta_i}{\sum_{i=1}^K\delta_i}=\geq\frac{\lambda}{L}+o(1).
\end{gather}

Thus we have
\begin{equation}
    \frac{\Delta}{\delta}\geq\frac{\lambda}{L}+o(1).
\end{equation}
\end{proof}

\section{$\lambda$-Sparsity}\label{sec:lambda}
\begin{figure}
    \centering
    \includegraphics[width=1.0\linewidth]{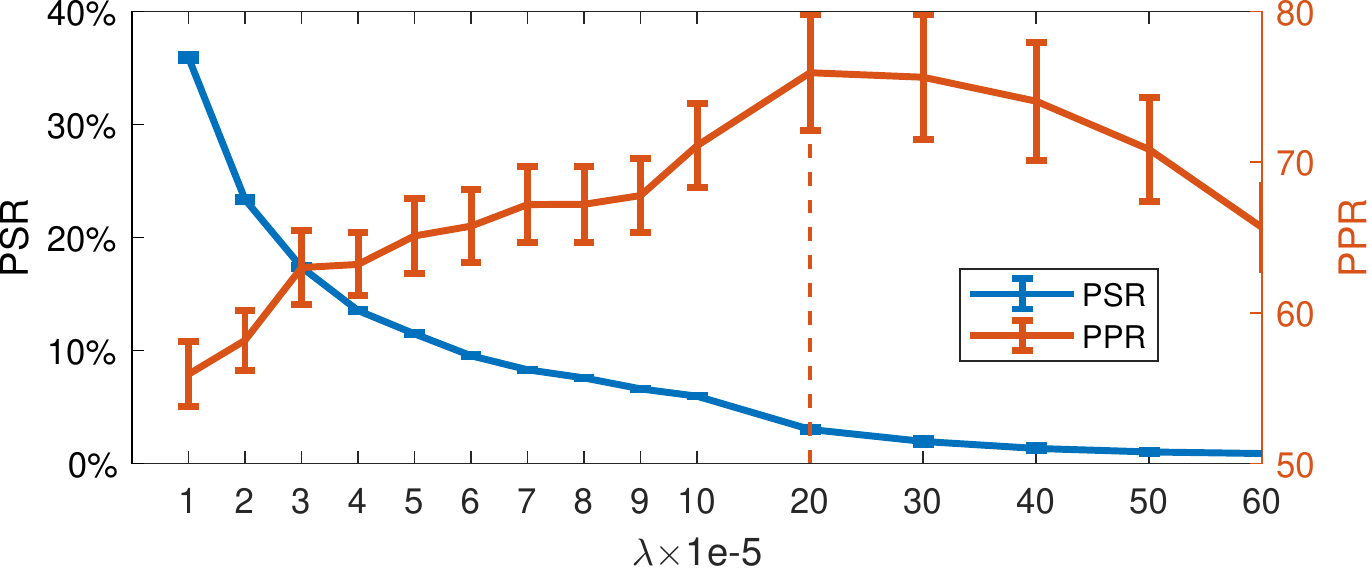}
    \caption{Pixel Principal Ratio (PPR) and Parameter Sparsity Ratio (PSR) of PKD under different $\lambda$. We find a turning point for the PPR metric, which also corresponds to the minimum volume of parameters to capture the knowledge of interest.}
    \label{fig:ppr_psr}
\end{figure}

For a given data domain and generator model, we find that the volume of parameters that are active to a knowledge of interest is an interesting property. To investigate it, we randomly sample 500 images in the StyleGAN2 generator of FFHQ domain, and conduct Dirac Knowledge Extrapolation to those data points with different hyper-parameter $\lambda$. We report the results of two metrics:  \textit{Pixel Principal Ratio} (PPR) and  \textit{Parameter Sparsity Ratio} (PSR). PPR measures the ratio of changes induced by PKD between the cross entropy $H(\mathbb{P}_{\Gtheta},\mathbb{P}_{\overline{l}})$ and the pixel value of the image, \ieno,
\begin{equation}
    PPR = \frac{\vert\log(\mathbb{P}_l(\Gtheta(\z)))-\log(\mathbb{P}_l(\GthetaX(\z)))\vert}{\frac{1}{hwc}\Vert\Gtheta(\z)-\GthetaX(\z)\Vert_2^2},
\end{equation}
where $h,w,c$ are height, width, and channel numbers of the image domain. PSR measures the ratio of parameters that correspond to non-zero updating during PKD. The remaining parameters can be viewed as inactive to the extrapolated knowledge. We report the mean values and standard deviations of these two measurements in Fig. \ref{fig:ppr_psr}. As indicated by Theorem \ref{th:lambda},  PPR monotonically increases before  $\lambda_{max}$ of 2e-4. It suggests that before $\lambda_{max}$, increasing $\lambda$ can further exclude redundant parameters in the PKD process. After $\lambda_{max}$, however, increasing $\lambda$ will also hurt the extraction of the knowledge of interest. On the other hand,  PSP monotonically decreases as $\lambda$ increases, which is the consequence of sparsity enforced by the $\ell_1$ penalty. Thus, the turning point of PPR should give the minimum ratio of parameters that are active to the knowledge of interest. Thus, in all the previous experiments, we set the hyper-parameter $\lambda$ to a value that is slightly smaller than $\lambda_{max}$ to secure better performance.

\section{Experiment Setting}\label{sec:exp_set}

\paragraph{Generative Model Choice} For FFHQ data domain, we use the pretrained StyleGAN2 generator offered by Awesome Pretrained StyleGAN2 \footnote{\url{https://github.com/justinpinkney/awesome-pretrained-stylegan2}} with config-f and $512\times 512$ resolution. For BreCaHAD data domain, we use the official pretrained StyleGAN2-ADA generator\footnote{\url{https://github.com/NVlabs/stylegan2-ada/}}. For ImageNet data domain, we use the BigGAN256-Deep model in the official TFhub repository \footnote{\url{https://tfhub.dev/deepmind/biggan-deep-256/1}}.
\paragraph{Posterior Estimation Model Choice} For FFHQ data domain, we use the official pretrained ResNet50 classifiers provided by StyleGAN2 authors \footnote{\url{https://github.com/NVlabs/stylegan2}} as the posterior distribution $\mathbb{P}_l$. For BreCaHAD, we train a ResNet50 regressive model on the annotation subset of BreCaHAD dataset. The annotations mark all cancer nucleus in the  tissue slice images. We train the ResNet50 regressive model to predict the number of cancer nucleus in each tissue slice image. We halt the training at the error rate of 10\% in the training set to avoid overfitting, as the training dataset is  small. The final predict scores are further normalized to $[0,1]$ to yield posterior estimation to the severity of cancer. For ImageNet data domain, we use the official ResNet50 classifier provided by TensorFlow \footnote{\url{https://www.tensorflow.org/api_docs/python/tf/keras/applications/resnet50/ResNet50}}. The ResNet50 classifier outputs a 1,000 dimensional vector, each of which predicts the posterior of a given image category. We choose the dimension corresponding to our knowledge of interest as the final posterior estimation $\mathbb{P}_l$.
\paragraph{Hyper-parameter Setting} For `Lipstick' extrapolation of FFHQ data domain, we set $K=4$; for `Gray Hair' extrapolation of FFHQ data domain, we set $K=7$. For all the other experiments, we set $K=10$. For FFHQ domain and BreCaHAD domain, we set $\epsilon=1e-3$; for ImageNet domain, we set $\epsilon=1e-5$. The choice of $\lambda$ is set according to Fig. \ref{fig:ppr_psr}, where we choose a $\lambda$ that is slightly smaller than $\lambda_{max}$ for each experiment. For FFHQ domain, we set $\lambda=1.8e-4$; for BreCaHAD domain, we set $\lambda=4.8e-4$; for ImageNet domain, we set $\lambda=1.3e-4$. For all Dirac Knowledge Extrapolations, we set $\xi=0.01$.
\paragraph{Training of InterFaceGAN} We obtain the semantic boundary vectors as directed by the InterFaceGAN paper. We use the ResNet50 classifiers provided by the StyleGAN2 authors to annotate 50,000 random samples of the StyleGAN2 generator, and then train a Support Vector Machine (SVM) to predict the binary annotations predicted by the ResNet50 classifiers. The normalized support vectors of those SVMs are chosen to serve as the semantic boundaries for latent image editing. 

\begin{figure*}
    \centering
    \includegraphics[width=1.0\linewidth]{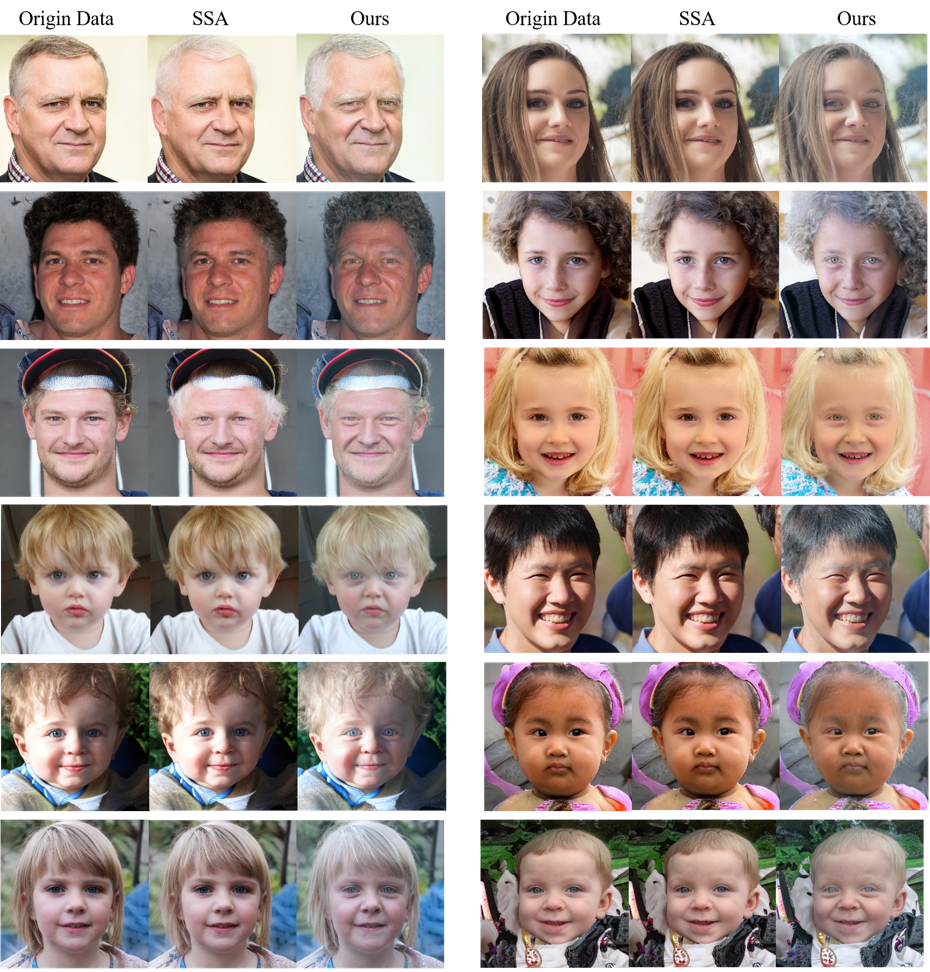}
    \caption{Extrapolation of knowledge `graph hair'. We compare our method with the latent image editing method Style Space Analysis (SSA). We set the editing strength of SSA to yield slightly stronger degree of hair grayness than ours in regular cases, \egno, middle-age male, as shown in the first three rows of the left side. We then investigate the performance of both methods in counterfactual cases---young women or children in gray hair. The results show that SSA can hardly handle the counterfactual cases, while our method still works well.}
    \label{fig:ssa}
\end{figure*}

\paragraph{Style Space Analysis \cite{wu2021stylespace}} We also compare the PKD method in Dirac distribution with another latent image editing method Style Space Analysis (SSA) \cite{wu2021stylespace}, the results are reported in Fig. \ref{fig:ssa}. We use the official code provided by Style Space Analysis authors \footnote{\url{https://github.com/betterze/StyleSpace}}.

\paragraph{User Study}
We conduct user studies on the FFHQ data domain for `mustache' extrapolation by recruiting 50 volunteers. We randomly sample 3,000 latent codes from the prior distribution $\mathcal{N}(\mathcal{O},\mathcal{I})$, and feed them to the pretrained StyleGAN2 generator to produce 3,000 synthesized facial images. The same operation is performed to the extrapolated distribution to yield another independent 3,000 synthesized facial images after knowledge extrapolation. Then we mix the StyleGAN2 synthesis facial images with the extrapolated synthesis facial images to yield a 6,000 testing set, and then randomly shuffle the testing set. All 50 volunteers are asked to answer the following questions for each image of the testing set: 1) Does the person in the image have mustache; 2) Is the person in the image a child without regard to mustache; 3) what is the gender of the person in the image without regard to mustache. For each question, the answer with the highest votes will be the winner.

\paragraph{Hardware Setting} To train StyleGAN2 on $512\times 512$ resolution FFHQ dataset and InterFaceGAN semantic boundaries, we use 8 NVIDIA V100 GPUs. To train the ResNet50 regressive model for BreCaHAD data domain, we use 1 NVIDIA GTX1080Ti GPU. For all the remaining experiments of knowledge extrapolation, we use 1 NVIDIA V100 GPU.

\begin{figure*}
    \centering
    \includegraphics[width=1.0\linewidth]{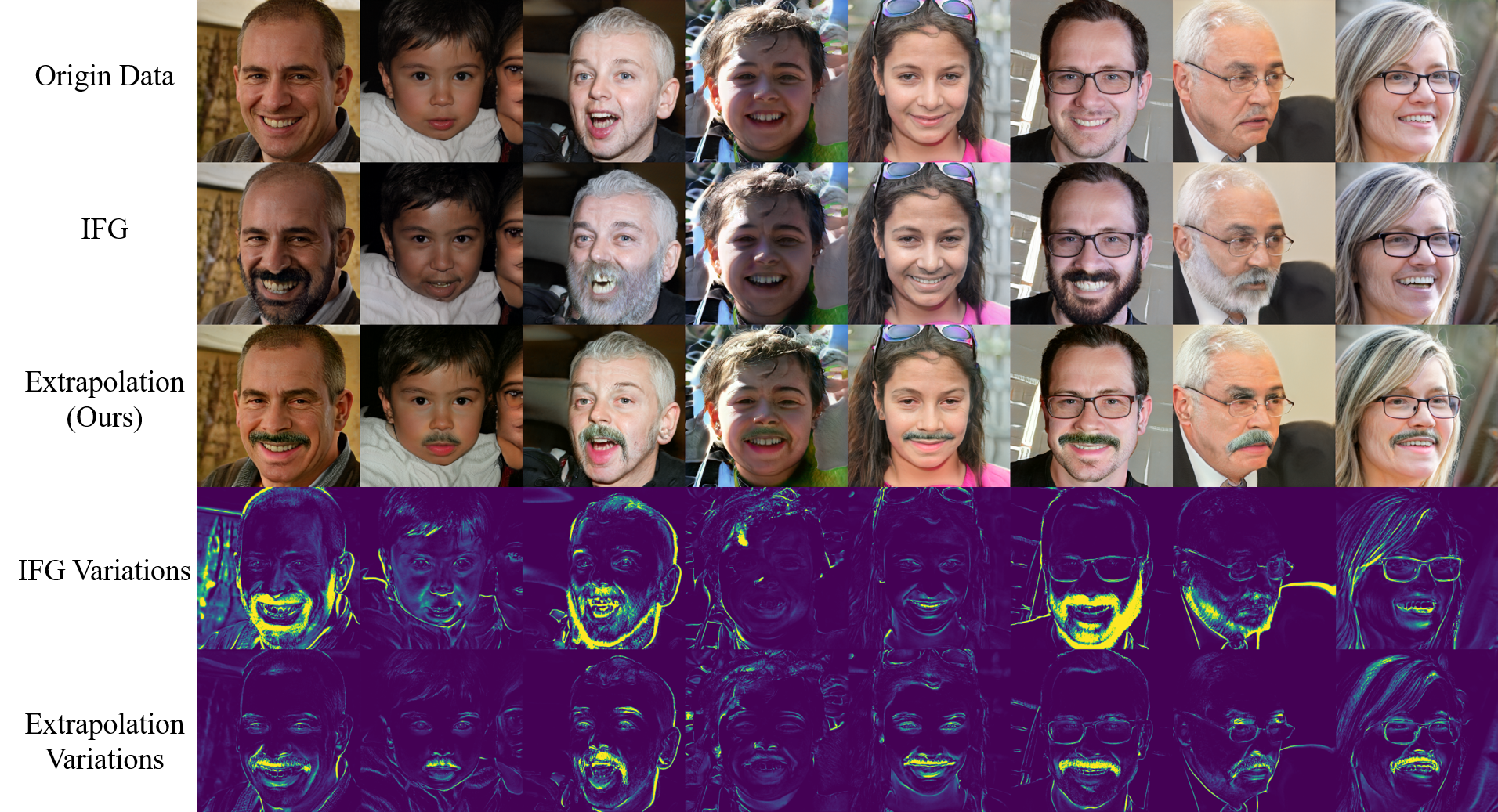}
    \caption{Dirac knowledge extrapolation of mustache on randomly sampled latent codes generating FFHQ faces. IFG denotes InterFaceGAN. IFG can successfully add mustache for regular cases, \egno, mature male, but fails the counterfactual cases.}
    \label{fig:ifgvpkd_unc}
\end{figure*}

\begin{figure*}
    \centering
    \includegraphics[width=1.0\linewidth]{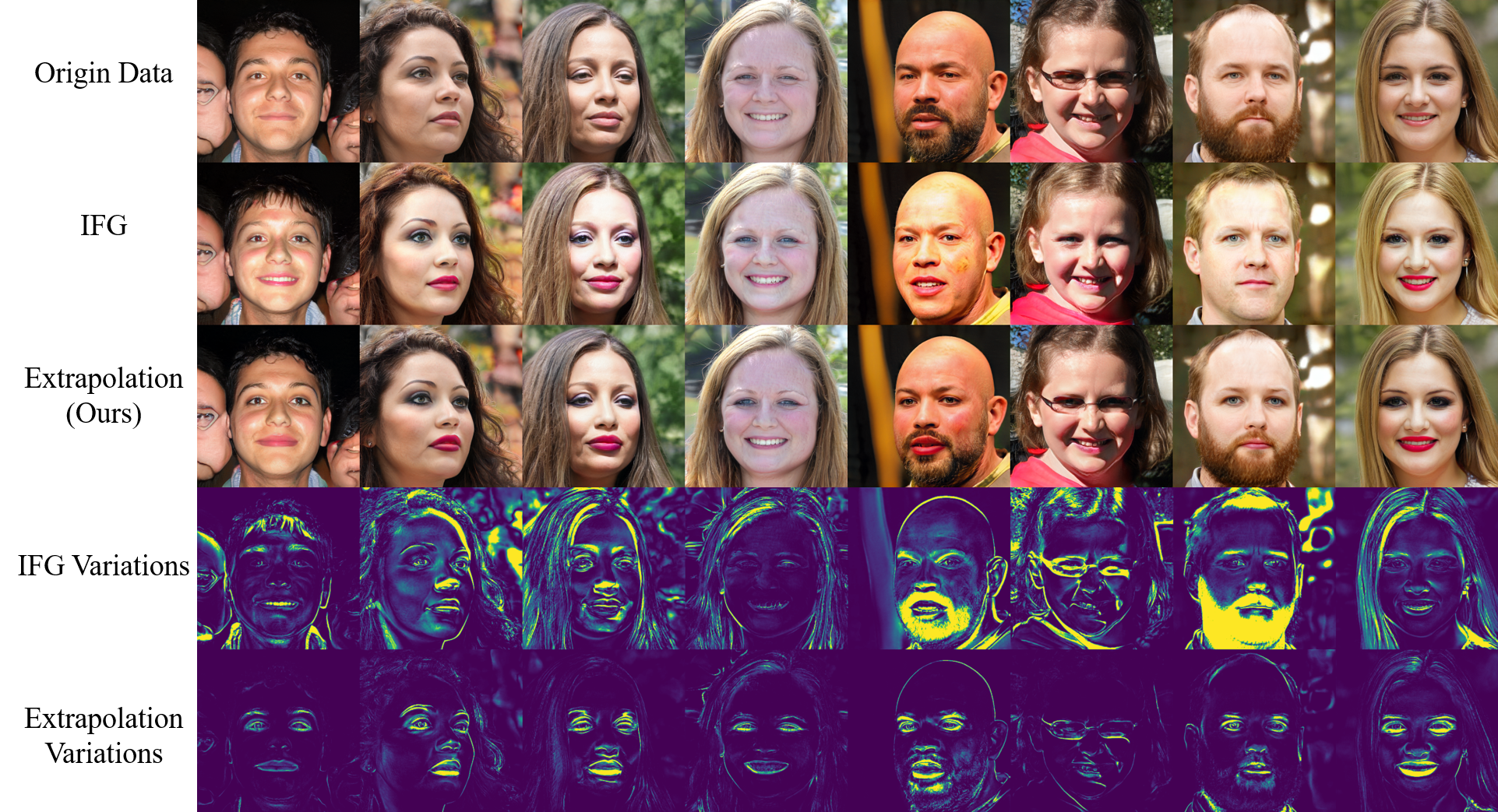}
    \caption{Dirac knowledge extrapolation of lipstick on randomly sampled latent codes generating FFHQ faces. IFG denotes InterFaceGAN. IFG can successfully add lipstick for regular cases, \egno, female, but fails the counterfactual cases, \egno, removing mustache or changing gender of the male cases.}
    \label{fig:ifgvpkd_unc2}
\end{figure*}
\begin{figure*}
    \centering
    \includegraphics[width=1.0\linewidth]{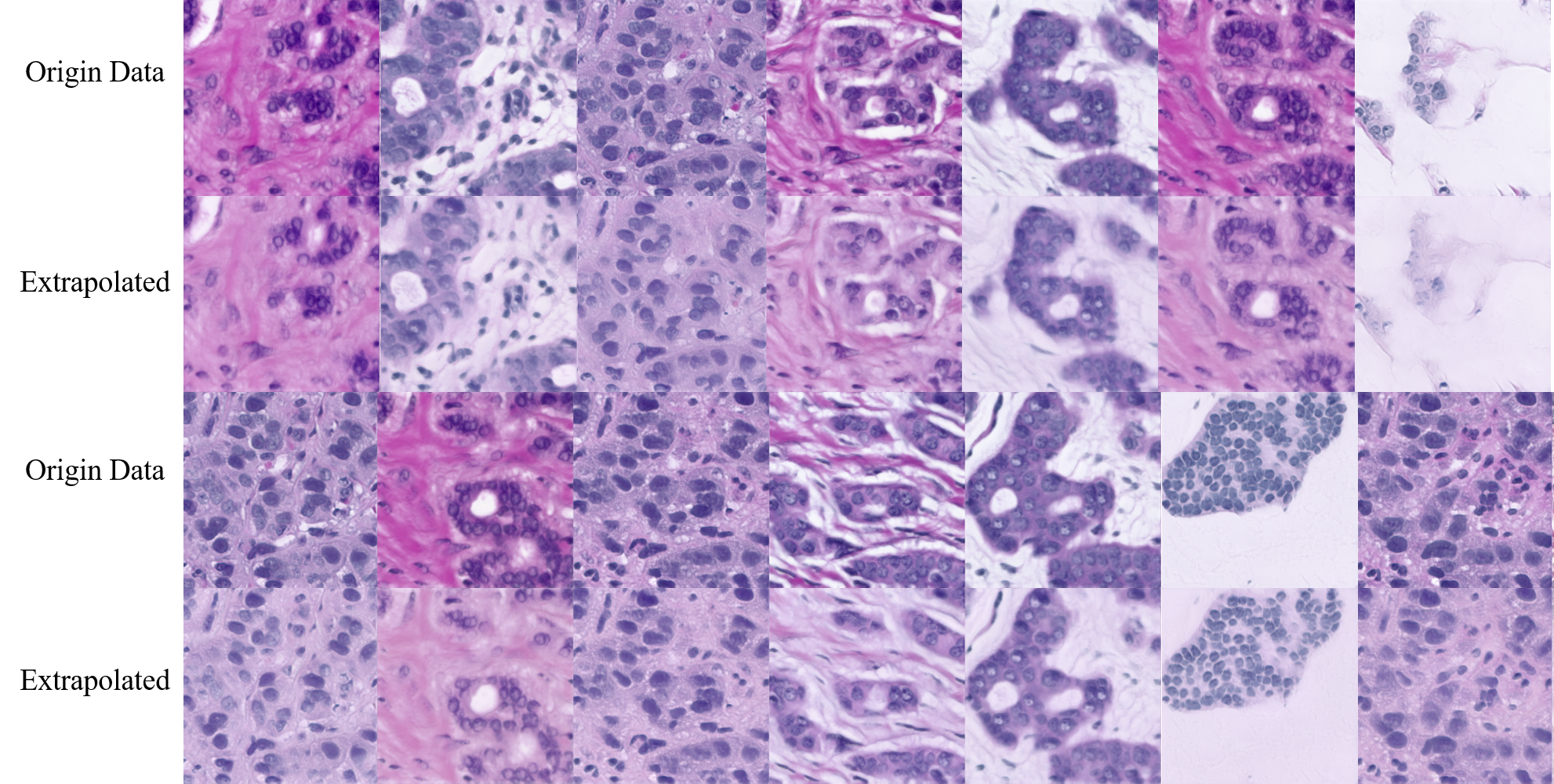}
    \caption{Dirac knowledge extrapolation of tissue slices of cancer patients. The cancer severity is reduced in the extrapolated cases.}
    \label{fig:my_ext_tumor2}
\end{figure*}

\begin{figure*}
    \centering
    \includegraphics[width=1.0\linewidth]{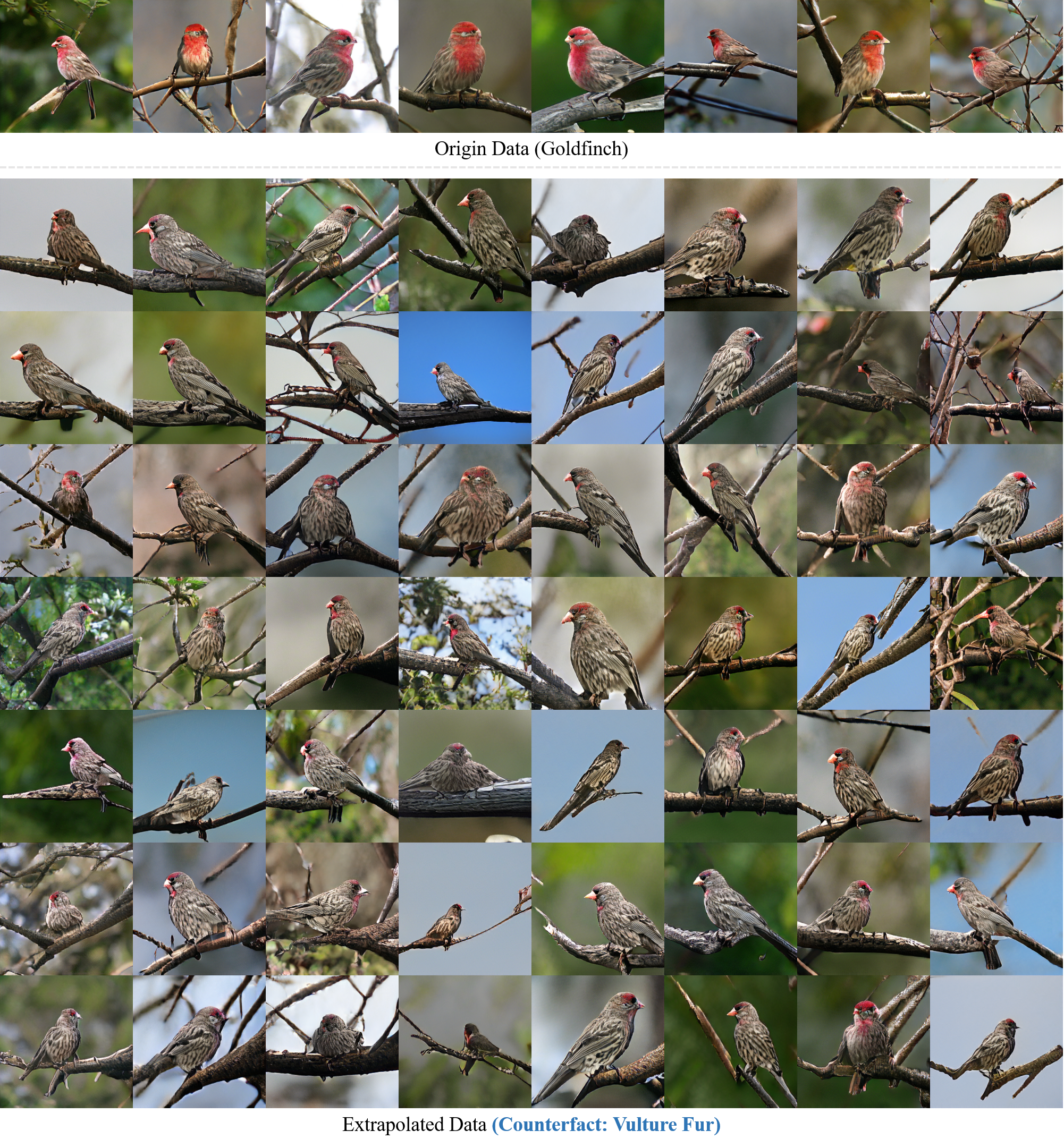}
    \caption{Uncurated lists of BigGAN generations after extrapolating knowledge dimensions.}
    \label{fig:biggan_23}
\end{figure*}

\end{document}